\documentclass[conference]{IEEEtran}
\usepackage{fancyhdr}

\usepackage{amsmath}
\usepackage{amssymb}
\usepackage{xcolor}
\usepackage{algorithm}
\usepackage{algpseudocode}

\let\oldReturn\Return
\renewcommand{\Return}{\State\oldReturn}
\usepackage{xspace}
\DeclareRobustCommand{\RELATE}{\textsf{ReLATE}\xspace}
\usepackage{graphicx}
\usepackage{pgfplots}
\usepackage{multirow}
\pgfplotsset{compat=1.16}
\usepackage{tikz}
\usetikzlibrary{patterns}
\usetikzlibrary{shapes}
\usepackage[nolist]{acronym}
\usepackage{paralist}
\usepackage{enumitem}
\usepackage{balance}
\usepackage{comment}
\usepackage{booktabs}
\usepackage{url}
\usepackage{cite}
\usepackage{hyperref}

\providecommand{\Description}[2][]{}

\definecolor{greyback}{RGB}{248,248,248}
\definecolor{deepblue}{RGB}{43,131,186}
\definecolor{greenalt}{RGB}{35,139,69}
\definecolor{darkred}{RGB}{215,25,28}
\definecolor{darkorange}{RGB}{253,174,97}
\definecolor{light-gray}{gray}{0.85}
\definecolor{lighter-gray}{gray}{0.95}
\definecolor{applegreen}{rgb}{0.55, 0.71, 0.0}
\definecolor{asparagus}{rgb}{0.53, 0.66, 0.42}
\definecolor{babyblueeyes}{rgb}{0.63, 0.79, 0.95}
\definecolor{burntsienna}{rgb}{0.91, 0.45, 0.32}
\definecolor{deepcarmine}{rgb}{0.66, 0.13, 0.24}
\definecolor{lightcornflowerblue}{rgb}{0.6, 0.81, 0.93}
\definecolor{steelblue}{rgb}{0.27, 0.51, 0.71}
\definecolor{pastelblue}{rgb}{0.68, 0.78, 0.81}
\definecolor{pastelbrown}{rgb}{0.51, 0.41, 0.33}
\definecolor{pastelgray}{rgb}{0.81, 0.81, 0.77}
\definecolor{pastelgreen}{RGB}{184,216,190}
\definecolor{pastelmagenta}{rgb}{0.96, 0.6, 0.76}
\definecolor{pastelorange}{rgb}{1.0, 0.7, 0.28}
\definecolor{pastelpink}{rgb}{1.0, 0.82, 0.86}
\definecolor{pastelpurple}{rgb}{0.7, 0.62, 0.71}
\definecolor{pastelred}{rgb}{1.0, 0.41, 0.38}
\definecolor{pastelviolet}{rgb}{0.8, 0.6, 0.79}
\definecolor{pastelyellow}{rgb}{0.99, 0.99, 0.59}
\definecolor{darkblue}{rgb}{0.0, 0.0, 0.55}
\definecolor{darkgray}{rgb}{0.675, 0.655, 0.60}
\definecolor{light-gray}{gray}{0.85}
\definecolor{lighter-gray}{gray}{0.95}
\definecolor{lighter-gray2}{gray}{0.98}

\newcommand{\VECELEM}[2]{${#1}_{#2}$}
\newcommand{\TENSOR}[1]{$\mathcal{#1}$}

\newcommand{\REALTWO}[2]{$\mathbb{R}^{#1\times #2}$}
\newcommand{\REALTHREE}[3]{$\mathbb{R}^{#1\times #2 \times #3}$}

\newcommand{\IGNORE}[1]{}

\newcommand{\GHZ}{\mbox{GHz}}

\newcommand{\MiB}{\mbox{MiB}}

\begin{document}

\title{ReLATE: Accelerating Tensor Decomposition via Safe and Efficient Learning of Sparse Encodings}

\author{
\IEEEauthorblockN{
Ahmed E. Helal\IEEEauthorrefmark{1},
Fabio Checconi\IEEEauthorrefmark{1},
Jan Laukemann\IEEEauthorrefmark{2},
Yongseok Soh\IEEEauthorrefmark{3},\\
Jesmin Jahan Tithi\IEEEauthorrefmark{1},
Fabrizio Petrini\IEEEauthorrefmark{1},
and Jee W. Choi\IEEEauthorrefmark{3}
}
\IEEEauthorblockA{
\IEEEauthorrefmark{1}\textit{Intel Corporation}, Santa Clara, CA, USA\\
\{ahmed.helal,fabio.checconi,jesmin.jahan.tithi,fabrizio.petrini\}@intel.com
}
\IEEEauthorblockA{
\IEEEauthorrefmark{2}\textit{University of Erlangen-Nürnberg}, Erlangen, Germany\\
jan.laukemann@fau.de
}
\IEEEauthorblockA{
\IEEEauthorrefmark{3}\textit{University of Oregon}, Eugene, OR, USA\\
\{ysoh,jeec\}@uoregon.edu
}
}

\maketitle

\thispagestyle{fancy}
\lhead{}
\rhead{}
\chead{}
\lfoot{\footnotesize{
SC26, November 15-20, 2026, Chicago, Illinois, USA
\newline 979-8-3195-4789-7/26/\$31.00 \copyright 2026 IEEE}}
\rfoot{}
\cfoot{}
\renewcommand{\headrulewidth}{0pt}
\renewcommand{\footrulewidth}{0pt}

\begin{abstract}
Tensor decomposition (TD) is essential for analyzing high-dimensional sparse data, yet its irregular computations and memory-access patterns pose major performance challenges on modern parallel processors. Prior works rely on expert-designed sparse tensor formats that fail to adapt to irregular tensor shapes and data distributions. We present the reinforcement-learned adaptive tensor encoding (\RELATE) framework, a learning-augmented method that discovers safe and efficient sparse encodings, without labeled examples, via a hybrid model-free and model-based algorithm that learns from both real and imagined actions. 
Moreover, \RELATE introduces elastic training, rule-driven action masking, and dynamics-informed action filtering to ensure correct encoding with bounded execution time, even during early learning. 
After offline training, with geometric-mean overhead of only $5.82\%$ relative to TD workflow time,
\RELATE deploys the best encoding with zero inference overhead.
Across diverse real-world sparse tensors, \RELATE consistently outperforms the best expert-designed 
format by up to $2 \times$, with a geometric-mean speedup of $1.38$--$1.41 \times$.
\end{abstract}

\begin{IEEEkeywords}
Sparse tensors, tensor decomposition, MTTKRP, parallel algorithms, multicore processing, reinforcement learning
\end{IEEEkeywords}

\section{Introduction}
Tensor decomposition (TD) generalizes principal component analysis to break down complex, high-dimensional sparse data, such as financial transactions, electronic health records, and user ratings~\cite{kobayashi2018extracting, yadav2018mining, symeonidis2016matrix}, into simpler, low-dimensional components that reveal the latent relationships and trends within the data. Although TD algorithms play a critical role in dimensionality reduction, compression, and analysis of multi-way data~\cite{KoBa09, sidiropoulos2017tensor, panagakis2021tensor, liu2023tensor}, they remain challenging to scale on modern parallel processors because of their highly irregular computations and memory-access patterns~\cite{choi2018blocking, alto_2021}. 
In contrast to dense tensors, 
sparse tensors can be represented in a multitude of ways, each tailored for different sparsity patterns and exhibiting varying storage and performance trade-offs~\cite{smith2017accelerating, laukemann2025accelerating}.
Moreover, TD algorithms perform computations across all dimensions (i.e., modes) of a sparse tensor, and even if a sparse representation is efficient for a particular mode, it might deliver subpar performance in other modes~\cite{smith2017accelerating, alto_2021}.

Since TD problems are both high-dimensional and data-dependent, their optimization space is vast (as detailed in $\S$\ref{sec:approach-ac}). However, state-of-the-art approaches rely on suboptimal heuristics for data representation and parallel execution that fail to account for the irregular shapes and data distributions of sparse tensors. Prior work improved TD performance using compressed sparse tensor representations, including the mode-specific compressed sparse fiber (CSF) format~\cite{Smith2015, smith2017accelerating} and the mode-agnostic hierarchical coordinate (HiCOO) format~\cite{li2018hicoo, li2019efficient}. While these sparse representations substantially outperform the de facto coordinate~(COO) format, they are oblivious to the highly irregular shapes and data distributions of sparse tensors~\cite{alto_2021, laukemann2025accelerating}. The state-of-the-art linearized tensor formats~\cite{alto_2021, blco_2022, alto_stream_2023, laukemann2025accelerating}, such as the mode-agnostic adaptive linearized tensor order (ALTO) format, tailor their tensor representation to the irregular shapes of sparse tensors, yet they overlook the underlying data distributions.
The sparse tensor format selection (SpTFS) framework~\cite{sun2020sptfs, sun2021input} uses supervised machine learning to predict the best among COO, HiCOO, and CSF formats to decompose a given sparse tensor. However, due to the lack of large-scale sparse tensor training sets and because the optimal solution (or ground truth) 
for such irregular workloads is unknown, supervised learning methods are impractical. 
Additionally, the performance of these methods is bounded by the best existing formats, even if they achieved oracle-level prediction accuracy. 

In place of suboptimal, input-oblivious heuristics, we propose the reinforcement-learned adaptive tensor encoding (\RELATE) framework. \RELATE follows the emerging learned or learning-augmented paradigm~\cite{mitzenmacher2022algorithms}, combining domain expertise with a rich problem formulation to navigate the solution space effectively.
Unlike classical auto-tuning, our learned algorithm goes beyond adjusting program parameters to construct an efficient sparse tensor encoding and the corresponding schedule of tensor compute and memory operations.  

Central to \RELATE is an autonomous agent that leverages deep reinforcement learning (DRL) and domain knowledge to improve the performance of high-dimensional data analytics \textit{without requiring labeled training samples}. Specifically, the agent learns the best sparse tensor encoding directly from interactions with the TD environment by taking actions and observing the environment state and reward. Importantly, \RELATE introduces \textit{a hybrid model-free and model-based} learning algorithm, where the agent employs an adaptive neural network that scales with environment complexity and learns from both real (evaluated) and imagined (estimated) actions to effectively handle large-scale tensors. 
\RELATE guarantees valid encodings and accelerates learning through \textit{rule-driven} action masking and \textit{dynamics-informed} action filtering to prune invalid actions during the early exploration stage and low-value actions in the delayed exploitation stage of the reinforcement learning process. Furthermore, \RELATE uses an elastic training span rather than a fixed episode budget, stopping early when reward improvements plateau under an exploration-dependent patience criterion.
To obtain reliable rewards in HPC environments that are both growing in complexity and prone to noise~\cite{tuncer2017diagnosing}, \RELATE uses a relative reward signal and adopts a decoupled server-client execution model that isolates learning and decision-making from reward evaluation by placing them on separate compute nodes.

By overcoming the limitations of prior supervised learning methods and tensor formats, \RELATE adapts to the irregular shapes and data distributions of sparse tensors and to the target hardware architectures. As a result, it discovers sparse representations that outperform expert-designed formats across a representative set of sparse tensors. In summary, the main contributions of this work are: 
\begin{itemize}[leftmargin=*]
\item We introduce \RELATE, a DRL framework that constructs efficient mode-agnostic sparse tensor representations without labeled examples. Unlike prior work~\cite{Smith2015, smith2017accelerating, li2018hicoo, alto_2021, laukemann2025accelerating}, \RELATE automatically adapts to both the asymmetric shapes and irregular data distributions of sparse tensors ($\S$\ref{sec:approach}).  
\item We tackle the exploration-exploitation dilemma by devising a domain-specific agent that effectively navigates the vast solution space of TD problems using elastic training, rule-driven action masking, and model-guided action filtering. Hence, \RELATE always generates safe (functionally correct with bounded execution time) sparse encodings,
even before learning the environment dynamics ($\S$\ref{sec:approach} and $\S$\ref{sec:results}).
\item On an Intel Emerald Rapids system, \RELATE delivers geometric-mean speedups of $1.38 \times$ and $1.41\times$ over the best expert format for original and randomly permuted sparse tensors, respectively, reaching up to $2 \times$ for large-scale, hyper-sparse tensors. Training incurs only $5.82\%$ geometric-mean overhead relative to TD workflow time ($\S$\ref{sec:results}). 
\end{itemize}

\section{Background and Related Work}
\subsection{Tensor Notations}
A tensor is an $N$-dimensional array, with each element addressed by an $N$-tuple index~$\textbf{i}~=~(i_1, i_2, \ldots, i_N)$.
The coordinate $i_n$ locates a tensor element along the $n^{\mathrm{th}}$ 
dimension or mode, whose length is $I_{n}$, with $n \in \{1, 2, \ldots, N\}$ and $i_n \in \{0, 1, \ldots, I_n - 1\}$.
A \emph{dense} $N$-dimensional (or a mode-$N$) tensor has $\prod_{n=1}^N I_n$ elements.
A tensor is sparse when most of its elements are zero.
We use the following notations:
\begin{enumerate}[leftmargin=*]
   \item Scalars are denoted by italicized lowercase letters (e.g., $c$).
   \item Vectors are denoted by bold lowercase letters (e.g., $\mathbf c$).
   \item Matrices are denoted by bold capital letters (e.g., $\mathbf C$). 
    \item Tensors are denoted by Euler script letters (e.g., $\mathcal X$).
    \item Fibers in a tensor generalize matrix rows and columns. 
    A mode-$n$ fiber of a tensor $\mathcal X$ is any vector obtained by fixing all indices of $\mathcal X$ \emph{except} the $n^{\mathrm{th}}$ index.
   \item To denote all elements along a given mode or dimension, we use the $:$ symbol.
   For example, $\mathbf{C}(n,:)$ indicates the $n^{\mathrm{th}}$ row of the matrix $\mathbf{C}$.
    \item 
    The mode-$n$ matricization of a tensor $\mathcal X$, denoted by $\mathbf X _{(n)}$, \emph{unfolds} $\mathcal X$ into a matrix by arranging its mode-$n$ fibers as the columns of $\mathbf X _{(n)}$.
    \item The Khatri-Rao product (KRP)~\cite{KR_product} is the column-wise Kronecker product of two matrices, denoted $\odot$. Given matrices 
	$\mathbf C^{(1)}$ $\in$ \REALTWO{I_{1}}{F} and $\mathbf C^{(2)}$ $\in$ \REALTWO{I_{2}}{F}, their KRP is $\mathbf K = \mathbf C^{(1)}\odot \mathbf C^{(2)} = \left[\textbf{c}_{1}^{(1)}\otimes \textbf{c}_{1}^{(2)}\ \textbf{c}_{2}^{(1)}\otimes \textbf{c}_{2}^{(2)} \dots \textbf{c}_{F}^{(1)}\otimes \textbf{c}_{F}^{(2)}\right]$, where $\otimes$ is the Kronecker product,  $\mathbf c_{f}$ is the $f^{\mathrm{th}}$ factor or column of the matrix $\mathbf C$, and $\mathbf K$ $\in$ \REALTWO{(I_{1}\cdot I_{2})}{F}.
   \item The Hadamard (element-wise) product is denoted by $*$.
\end{enumerate}

\subsection{Tensor Decomposition (TD)}\label{sec:background-TD}
\begin{figure}[tb]
\centering
\includegraphics[width=1.0\linewidth]{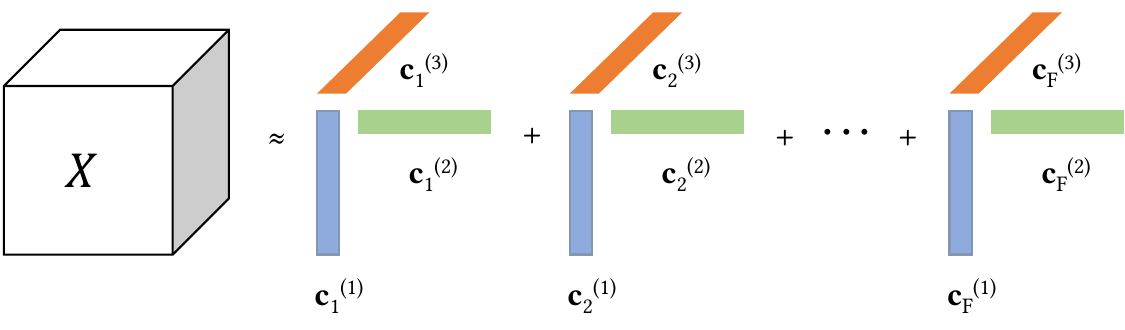}
    \caption{Canonical Polyadic Decomposition (CPD) of a mode-$3$ tensor $\mathcal X$.
    There are $F$ rank-one tensors (or factors) formed by the outer product of three vectors $\mathbf c_{f}^{(1)}$, $\mathbf c_{f}^{(2)}$, and $\mathbf c_{f}^{(3)}$, where $f \in \{1,2,\ldots,F\}$.
    Vectors along the same mode are grouped as columns of a \emph{factor matrix}.
    For example, $\mathbf c_{1}^{(1)}$, $\mathbf c_{2}^{(1)}$, \ldots, $\mathbf c_{F}^{(1)}$ constitute the columns of the mode-$1$ factor matrix $\mathbf C^{(1)}$.}
    \label{fig:CP}
\end{figure}

The survey by Kolda and Bader~\cite{KoBa09} provides a thorough overview of TD algorithms and applications. 
This work targets algorithms that iteratively compute the most popular TD model for \emph{sparse} tensors, the canonical polyadic decomposition (CPD), which approximates a tensor as a finite sum of rank‑one tensors, each representing a principal component or factor, as illustrated in Figure~\ref{fig:CP}. However, our ideas apply broadly to other tensor decomposition models and algorithms, due to the operation-agnostic nature of the framework.

\begin{figure}[tb]
\centering
\includegraphics[width=1.0\linewidth]{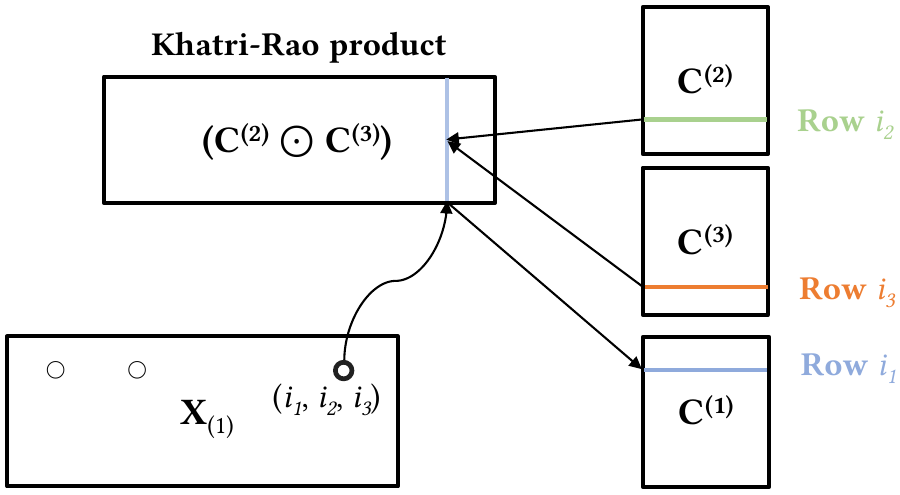}
    \caption{Dataflow of mode-$1$ MTTKRP on a mode-$3$ tensor $\mathcal X$. For every nonzero element at coordinates $(i_1, i_2, i_3)$, rows $i_2$ and $i_3$ are fetched from factor matrices $\mathbf{C}^{(2)}$ and $\mathbf{C}^{(3)}$, respectively. 
    Their element-wise product is computed to form the Khatri-Rao product, which is then scaled by the nonzero value and accumulated into row $i_1$ of factor matrix $\mathbf{C}^{(1)}$.
    }
    \label{fig:MTTKRP}
\end{figure}

The key performance bottleneck in the target TD algorithms is the matricized tensor times Khatri-Rao product (MTTKRP)~\cite{Smith2015, laukemann2025accelerating}, which must be executed along each mode in every iteration. The mode-$n$ MTTKRP on an $N$-dimensional tensor $\mathcal X$ is defined as
$\mathbf X _{(n)} \left(\bigodot_{k=1, k\neq n}^{N} \mathbf{C}^{(k)}\right)$, where $\mathbf C^{(n)}\in$ \REALTWO{I_{n}}{F} $\forall n\in \{1, 2, \ldots, N\}$ are the factor matrices. 
Computing MTTKRP requires iterating over every nonzero element in the sparse tensor and updating the corresponding row of the target factor matrix. For mode-$n$ MTTKRP, a nonzero element at index~$\textbf{i}~=~(i_1, i_2, \ldots, i_N)$ updates row $i_{n}$ of the $\mathbf C^{(n)}$ factor matrix after reading rows $i_{k}$ from $\mathbf{C}^{(k)}$ factor matrices $\forall k\in \{1, 2, \ldots, n-1, n+1, \ldots, N\}$, as depicted in Figure~\ref{fig:MTTKRP}. Consequently, MTTKRP is memory intensive and can incur substantial synchronization overhead when accumulating the partial results into factor matrices in parallel. Because finding a low-rank tensor approximation is NP-hard for tensors of order three or higher~\cite{de2008tensor}, real-world applications typically compute many approximations using TD before achieving a satisfactory fit. Each TD can take hundreds of iterations and executes MTTKRP along all $N$ modes in every iteration before converging to a solution~\cite{KoBa09, Smith2015, 10.1002/nla.2511}.   

Prior work focused on accelerating parallel MTTKRP using sparse tensor formats that can be classified as mode-specific or mode-agnostic. Mode-specific formats~\cite{Smith2015, Smith2015a, smith2017accelerating, nisa2019load, nisa2019efficient, kurt2022sparsity, kjolstad:2017:taco} generalize the classical compressed sparse row (CSR) format and typically require multiple tensor copies for best performance. Mode-agnostic formats~\cite{li2018hicoo, li2019efficient, alto_2021, blco_2022, alto_stream_2023, laukemann2025accelerating} maintain a single tensor copy by storing nonzero elements and their compressed coordinates in list- or block-based representations. Researchers applied supervised learning~\cite{sun2020sptfs, sun2021input} to predict the best mode-specific or mode-agnostic format for a given sparse tensor. 
Other approaches optimize sequences of MTTKRP operations by storing intermediate results for reuse~\cite{li2017model, kurt2022sparsity,kaya2018trees}.
Laukemann et al.~\cite{laukemann2025accelerating} provide a detailed review of sparse tensor formats and benchmark their performance across different TD algorithms and data types, showing that the main bottleneck is MTTKRP or an MTTKRP-like kernel.

\subsection{Deep Reinforcement Learning (DRL)}
Reinforcement learning (RL) is a goal-directed, decision-making process that emerged in biological brains to enable flexible behavioral responses to the environment~\cite{bennett2023brief}. DRL combines the trial‐and‐error learning methodology of RL with the automatic feature selection and hierarchical representation learning of modern deep neural networks~\cite{mnih2013playing, mnih2015human}. Contrary to a common misconception, trial-and-error learning is not random, as information about the target environment grows with every trial, especially when guided by a model capable of retaining favorable outcomes and discarding unfavorable ones. Specifically, a DRL-based agent learns a data-driven policy for a sequential decision-making problem by interacting with the environment through actions and observing the resulting state and reward or punishment signal. Formally, this Markov Decision Process (MDP) is defined by a tuple $\langle S, A, R, T, \gamma \rangle$, where $S$ is a set of environment states, $A$ is a set of actions, $R(s_t, a_t)$ is a reward function, $T(s_t, a_t, s_{t+1})$ is a transition function, and $\gamma \in [0, 1]$ is a discount factor. At each step or time instance $t$, the agent moves the environment from its current state $s_t \in S$ to a new state $s_{t+1} \in S$ by selecting an action $a_t \in A$, and then it receives a reward $r_t = R(s_t, a_t)$. An interaction (or observation) is thus defined by a tuple $\langle s_t, a_t, r_t, s_{t+1} \rangle$. A policy $\pi$ determines which action to take for each state $s_t \in S$. The goal is to learn the policy $\pi^{*}$ that maximizes the cumulative future reward, i.e., the sum of rewards $r_t$ discounted by a factor $\gamma$ at each time step $t$.    

One challenge in this learning approach is sample inefficiency: the agent typically requires many interactions with the environment to learn an effective policy. To improve sample efficiency, deep Q-network (DQN)~\cite{mnih2013playing, mnih2015human} uses an experience replay buffer to collect recent observations and learn by sampling minibatches from the buffer instead of only from the current experience, enabling \textit{offline} learning with bounded memory. 
Additionally, this replay mechanism enables DQN to 
mitigate correlation bias between consecutive observations, which deviates from the i.i.d. assumption of gradient-based optimizers.
Specifically, if the value $Q(s_t, a_t)$ is the future reward of a state-action pair, the optimal policy $\pi^* = \operatorname{argmax}_{a_t \in A} Q^{*}(s_t, a_t)$ selects the highest-valued action in all states. DQN approximates the value function $Q(s_t, a_t)$ with a deep neural policy network that estimates the values of all actions in a given state. To improve learning stability, double DQN~\cite{van2016deep} separates action selection, using a policy network, from action evaluation, with a target network.
Instead of random sampling from the replay buffer, prioritized replay~\cite{schaul2015prioritized} samples important experiences more frequently to further improve and accelerate learning~\cite{hessel2018rainbow}.

DRL has recently been applied to many performance-centric problems, notably in code generation~\cite{mankowitz2023faster, fawzi2022discovering}, compiler optimization~\cite{mammadli2020static, ahn2019reinforcement}, and job scheduling~\cite{gu2025deep, mangalampalli2024efficient}. In the domain of sparse matrix and tensor formats, researchers have applied DRL to automate format selection and tuning~\cite{armstrong2008reinforcement, chen2023accelerating, 10818187}, yet most prior work formulates these tasks as classification or regression problems and uses supervised learning~\cite{sun2020sptfs, sun2021input, zhao2018bridging, xie2019ia, stylianou2023optimizing, xiao2024machine, yesil2022dense, gao2024revisiting}.

\section{The ReLATE Framework}\label{sec:approach}

\subsection{Motivation and Challenges}\label{sec:approach-ch}
Devising a learnable sparse tensor format is challenging because of both the vast design space of this high-dimensional, data-dependent problem and the overhead and complexity of reward evaluation. Thus, it can be computationally prohibitive for DRL agents to explore such a massive state-action space and learn effective encoding policies from a noisy, high-latency reward signal~\cite{mnih2015human}. 

Moreover, using existing sparse formats as a basis for learned sparse tensor encodings is infeasible because they yield state–action spaces that are either too small or excessively large. For instance, the mode-specific CSF format~\cite{Smith2015, smith2017accelerating} splits the tensor into slices and fibers along a mode order, typically requiring one tensor copy per mode to perform well~\cite{smith2017accelerating, laukemann2025accelerating}. 
Since the performance of CSF largely depends on the choice of the root mode, with the ordering of the remaining modes having limited effect, a CSF-based format exposes only a few meaningful options to the learner. Applying a learning-augmented approach would therefore require a fundamental redesign of CSF to surface a richer set of construction decisions.
While block-based formats~\cite{li2018hicoo, li2019efficient} are mode-agnostic and only keep one tensor copy, using block size as the learnable parameter leads to a large state-action space proportional to the mode lengths.
Furthermore, unlike sparse matrices, block size typically has limited performance impact on sparse tensors, which exhibit unstructured and extreme sparsity due to the curse of dimensionality~\cite{alto_2021, laukemann2025accelerating}. Similarly, the number of ways to represent a sparse tensor using linearized tensor formats~\cite{alto_2021, blco_2022, laukemann2025accelerating} 
grows with both the mode lengths and the number of modes.

Even with a tractable state–action space, reward evaluation is expensive because it requires executing TD kernels on the target hardware. Additionally, modern HPC systems suffer from high run-time variability, driven by hardware complexity, resource contention, and operating system jitter~\cite{tuncer2017diagnosing}, which can significantly affect reward accuracy and DRL convergence. 

\subsection{ReLATE Overview}

\begin{figure}[tb]
\centering
\includegraphics[width=1.0\linewidth]{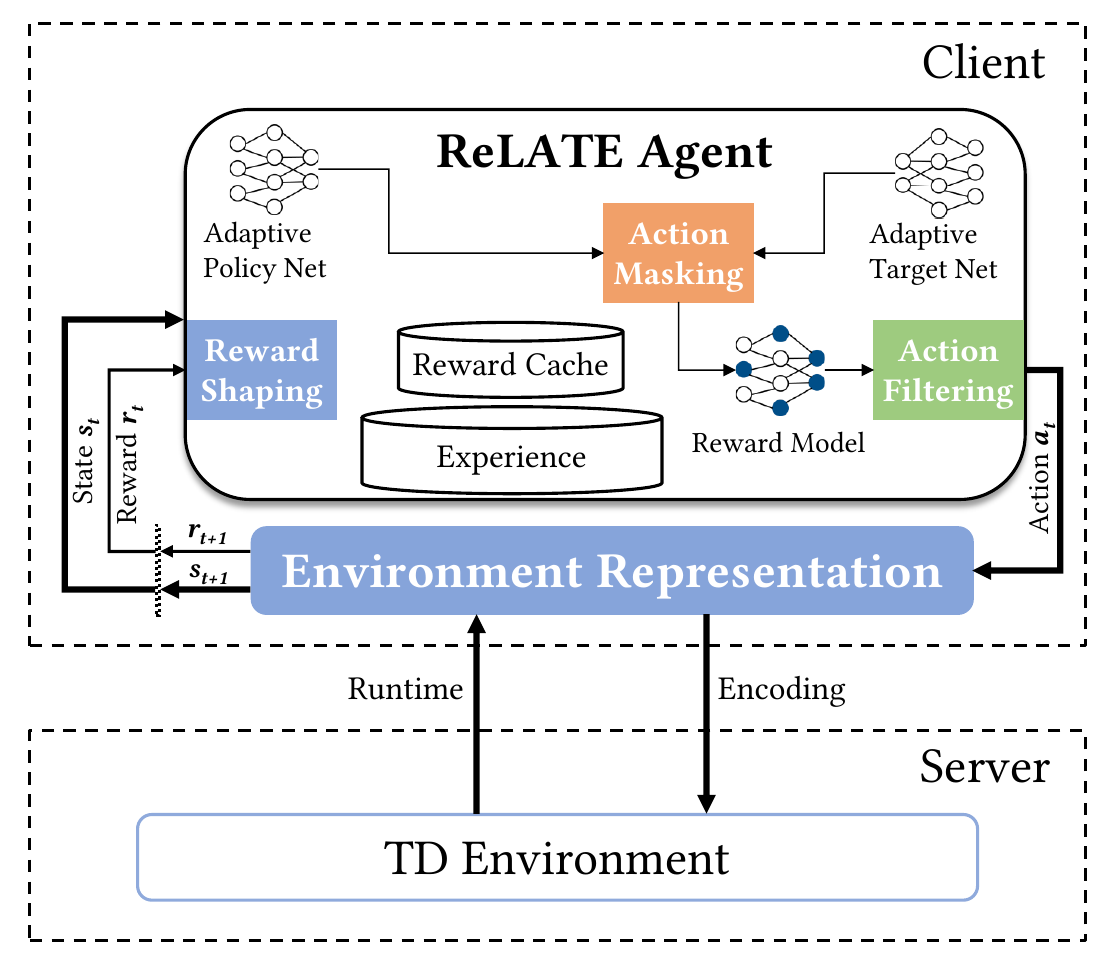}
	\caption{Overview of the \RELATE framework for learning efficient sparse tensor representations through direct interactions with the TD environment. The agent maintains policy and target networks, applies rule-driven action masking, and constructs tensor encodings one action at a time. Terminal encodings are sent to the environment for evaluation. The resulting rewards are used to train the policy and reward model, while reward caching and action filtering avoid unnecessary benchmark evaluations.}
\label{fig:relate}
\end{figure}

Figure~\ref{fig:relate} shows the high-level organization of \RELATE, a DRL-based framework for encoding sparse tensors using off-policy learning and a decoupled server-client execution model.
By separating sample collection (experience) from 
the policy being learned, \RELATE enables offline learning and improves sample efficiency. When the client (agent) and server (environment) are deployed on different compute nodes, \RELATE isolates learning from reward evaluation to reduce environment noise and improve reward accuracy. In addition, this decoupled execution allows \RELATE to utilize idle data center capacity to improve TD performance at minimal cost.

Specifically, sparse tensor encoding, together with the TD algorithm, determines the schedule of compute and memory operations, and in turn the parallel performance.
Our learned encoding leverages the state-of-the-art ALTO format~\cite{alto_2021, laukemann2025accelerating} because of its mode-agnostic nature and superior performance over prior formats. However, as explained in $\S$\ref{sec:approach-prob}, linearized tensor formats have an intractable state-action space that is combinatorial in the number of tensor modes and mode lengths. 
Thus, \RELATE devises an environment representation that reduces the state-action space without severely restricting the agent.
Furthermore, it models the search for the best sparse tensor encoding as a sequential decision problem, where the DRL agent transitions the environment from its initial state to a terminal state through a series of actions. 
In this model, the terminal state represents an encoding of the target sparse tensor, and its reward is computed as speedup over ALTO.
By interacting with the TD environment and collecting complete trajectories (i.e., sequences of observations ending in terminal states) along with their rewards, the agent learns a policy that maximizes the overall reward (i.e., speedup).

In this problem formulation, there are two 
main bottlenecks: training cost and reward evaluation overhead. Training cost is proportional to the state-action space, which depends on the tensor shape, i.e., the number of dimensions and their lengths. Reward evaluation overhead depends not only on the tensor shape but also on the number of nonzero elements. To tackle complex, high-dimensional environments with extensive dimension lengths, \RELATE introduces novel mechanisms to make learning tractable.
Instead of using negative rewards (punishments), \RELATE accelerates learning by using action masking to ensure that every episode or trajectory ends with a valid tensor encoding. 
Thus, the agent learns only from valid episodes and receives a normalized reward at the end of every episode. In this
daunting environment with sparse, delayed rewards, the normalized (differential) reward signal not only improves learning stability but also allows the agent to distinguish promising trajectories from the multitude of possible ones. 
To improve convergence, credit from the delayed reward is assigned back to earlier actions through reward shaping~\cite{10.5555/645528.657613}.
Importantly, \RELATE uses an elastic training span that adapts training duration to learning progress, stopping early when rewards plateau over a patience window proportional to the agent’s exploration rate.
Additionally, \RELATE employs convolutional neural networks (CNNs) to encode the hierarchical spatial information of the target environment and uses an adaptive neural network architecture, where the number of hidden units scales with the size of the state-action space.     

Due to the overhead of reward evaluation, \RELATE keeps a reward cache to avoid redundant evaluation of known valid encodings. Moreover, it introduces an action selection and filtering algorithm that allows early exploration of new actions and delayed exploitation of the agent’s learned knowledge.
That is, \RELATE uses a hybrid model-free and model-based reinforcement learning approach, in which the agent progressively builds a reward model of the environment from its exploration experience. Once the accuracy of the reward model is high (e.g., more than 90\%), the agent switches to learning from both real actions (reward evaluation) and imagined actions (reward model). Hence, if an imagined action is deemed high-value, 
in terms of its potential reward, 
the agent evaluates the actual reward and uses it to further refine the reward model. 
This hybrid learning enables safe and fast navigation of the vast solution space by reducing real slowdown (compared to the expert format) encountered during the exploration stage and by accelerating the reward evaluation during the exploitation stage.

In summary, the proposed mechanisms in \RELATE serve distinct roles: action masking guarantees encoding validity by construction; action filtering excludes candidate encodings predicted to yield low rewards; reward caching reuses exact measurements from previously evaluated encodings; and elastic early stopping concludes training when reward improvements plateau.

\subsection{Problem Formulation}\label{sec:approach-prob}
As discussed in $\S$\ref{sec:approach-ch}, prior sparse tensor formats are ill-suited for goal-directed learning because they are restricted to structured sparsity and/or have state–action spaces that are either overly constrained or unmanageably large. \RELATE learns efficient sparse tensor representations by reformulating this hard problem into a tractable one, mapping the construction of a linearized tensor format into a Markov Decision Process~(MDP) amenable to DRL-based methods. 

\begin{figure}[tb]
\centering
\includegraphics[width=1.0\linewidth]{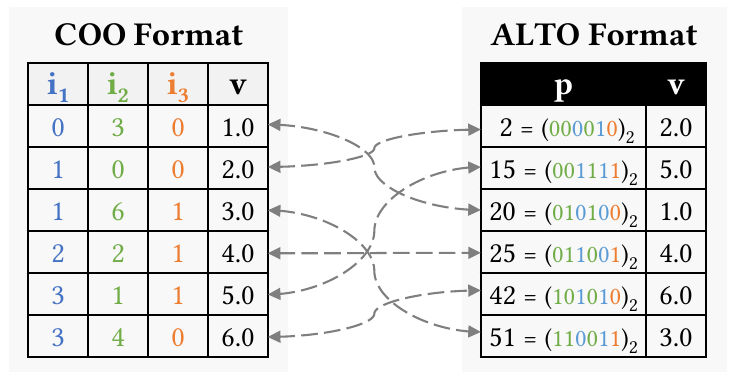}
	\caption{A linearized sparse tensor representation based on the ALTO format for a $4\times 8\times 2$ tensor with six nonzero elements. ALTO generates a compact mode-agnostic encoding amenable to efficient memory access and parallel execution by interleaving bits from the $N$-tuple index or coordinates ($i_1$, $i_2$, $i_3$) of nonzero elements.}
\label{fig:alto}
\end{figure}

State-of-the-art linearized tensor formats~\cite{blco_2022, laukemann2025accelerating} impose a total locality-preserving, mode-agnostic order on the nonzero elements of a sparse tensor to enable compact storage coupled with fast tensor access and traversal. 
Formally, a linearized sparse tensor \TENSOR{X} $= \{$\VECELEM{x}{1}, \VECELEM{x}{2}, $\dots$, \VECELEM{x}{M}$\}$ is an ordered collection of nonzero elements, where each element \VECELEM{x}{i} $= \langle v_{i}, p_{i} \rangle $ has a value $v_{i}$ and a position $p_{i}$~\cite{laukemann2025accelerating}. The position $p_{i}$ of the $i^{\mathrm{th}}$ nonzero element is a compact linear encoding of the $N$-tuple index, or coordinates, $\textbf{i}~=(i_{1},i_{2},\dots,i_{N})$, with $i_n \in \{0, 1, \ldots, I_n-1\}$ and $n \in \{1, 2, \ldots, N\}$. Precisely, the linear encoding $p$ uses $\ell(p) = \sum_{n=1}^{N} \ell(n) \text{ bits}$, with $\ell(n) = \lceil \log_2 I_{n} \rceil $, to represent the coordinates of nonzero elements as a sequence of bits $(b^{\ell(p) - 1} \dots b^{1} ~b^{0})_{2}$, where $b^{j}$ denotes the $(j+1)^{\mathrm{th}}$ bit of that encoding.
The coordinates $\textbf{i}$ of a nonzero element can be quickly derived from its position $p_{i}$, and vice versa, using simple bit-level operations that can be overlapped with memory-intensive tensor computations~\cite{alto_2021, blco_2022}. 
In contrast, traditional space-filling curves~\cite{peano1890courbe}, which typically target dense data, use $N \times \max_{n=1}^{N} \ell(n) \text{ bits}$ for linear indexing~\cite{alto_2021}.
This can be much more expensive because real-world tensors generally have skewed shapes (see Table~\ref{tab:problems}).
The linearized ALTO format and its variants substantially reduce the indexing size using a non-fractal encoding scheme that adapts to the irregular shapes of sparse tensors. This encoding partitions the tensor space along the longest mode first, which is equivalent to interleaving bits from the $N$-dimensional coordinates, starting with the shortest mode~\cite{laukemann2025accelerating}. Figure~\ref{fig:alto} shows a $4\times 8\times 2$ sparse tensor with six nonzero elements encoded in the COO and ALTO formats~\cite{laukemann2025accelerating}. 

While the resulting one-dimensional layout of linearized formats accommodates the asymmetric shapes of sparse tensors, it ignores their nonuniform and highly skewed data distributions, which affect the data reuse, workload balance, and synchronization cost of parallel TD operations~\cite{alto_2021, laukemann2025accelerating}. Hence, efficient high-dimensional data analysis requires 
choosing a linearization scheme that matches the unique sparsity patterns of each tensor. However, because each bit of the linear 
encoding $p$ can, in general, be selected to represent any bit of the $N$-tuple index $\textbf{i}$, the number of possible representations grows rapidly with the number of modes and mode lengths. Assuming, without loss of generality, that all tensor modes have the same length, there are $(N\,\ell(n))!$ possible linear formats. For a small 4-dimensional tensor with a mode length that fits into a single byte, the number of possible linearized representations is astronomically large ($(4\times 8)! \approx 2.63 \times 10^{35}$). 

Since no single linearized format is optimal across all sparse tensor data sets, \RELATE bridges this gap by searching a subset of the nearly infinite ways to linearly arrange the multi-dimensional data points of a tensor. Hence, instead of exploring $(N\,\ell(n))!$ possible linearized formats, \RELATE considers $ (N\,\ell(n))! \big/ (\ell(n)!)^{N} $ 
interleaved linear encodings ($\approx 9.96\times10^{16}$ for the prior example) to uncover effective sparse representations that improve TD performance ($\S$\ref{sec:approach-ac}). 
By decomposing the combinatorial choice of a linearized tensor encoding into $N\,\ell(n) $ simple decisions, \RELATE transforms an intractable problem into an MDP defined by states, actions, transitions, and rewards that can be tackled by DRL agents.

\subsubsection{Environment States} 
\begin{figure}[tb]
\centering
\includegraphics[width=1.0\linewidth]{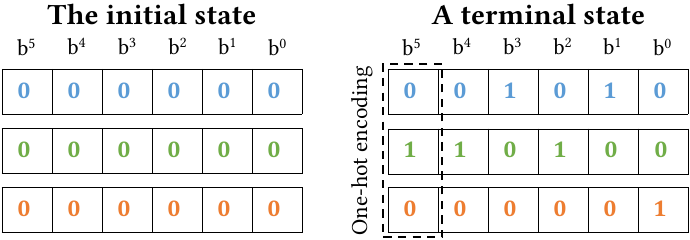}
	\caption{The environment representation of the example sparse tensor in Figure~\ref{fig:alto}. 
    Row $n$ in the encoding matrix represents tensor mode $n$, column $j$ represents bit $b^j$ of the linear encoding, and a hot entry at $[n,j]$ maps the next unused low-order bit of coordinate $i_n$ to $b^j$.
    The initial state indicates no encoding is selected, while the terminal state may correspond to any linearized sparse encoding, such as the ALTO encoding shown in the figure.}
\label{fig:env}
\end{figure}

\RELATE defines an encoding environment with three state types: initial, intermediate, and terminal. 
The agent drives the environment from its initial state (no encoding) to a terminal state (a linearized encoding) through a sequence of actions. A state $s_{t} \in S$ is an encoding matrix with $N$ rows and $\ell(p)$ columns that summarizes the bit-mapping decisions made so far. 
Each column $j$ can have at most one hot bit, indicating that the next low bit from a mode index $i_{n}$ (if the $n^{\mathrm{th}}$ bit in column $j$ is hot) is mapped to the $(j+1)^{\mathrm{th}}$ bit in the linear encoding (or bit $b^{j}$). 
Hence, in an intermediate state, the number of hot bits is strictly less than $\ell(p)$, while a terminal state has exactly $\ell(p)$ hot bits.
Figure~\ref{fig:env} illustrates this environment representation for the example sparse tensor from Figure~\ref{fig:alto}, with a terminal state representing the ALTO-based linearized encoding. In this example, the terminal state indicates that the first and second bits from the mode index or coordinate $i_{1}$ (i.e., $b_{i_{1}}^{0}$ and $b_{i_{1}}^{1}$) are mapped to the second and fourth bits in the linear encoding $p$ (i.e., $b^{1}$ and $b^{3}$), respectively.
Although a more compact representation using a vector of length $\ell(p)$ is possible, it requires integer rather than one-hot encoding for tensor modes, which introduces a bias and leads to poor learning performance by implying an arbitrary ordinal relationship across modes where none exists. Additionally, generalizing the environment representation so that each cell $[n, j]$ in the encoding matrix indicates which bit from a mode index $i_{n}$ is mapped to the $(j+1)^{\mathrm{th}}$ bit in the linear encoding $p$ leads to a prohibitively large ($\ell(p)!$) state space, as explained in $\S$\ref{sec:approach-ac}.      

\subsubsection{Actions}\label{sec:approach-ac}
In every episode, the agent performs $\ell(p)$ actions to transition the environment from its initial state to a terminal state by selecting the bits of the linear encoding $p$. In a step $t \in \{0, 1, \ldots, \ell(p) - 1\}$, the agent picks a mode index $i_{n}$ out of the $N$ tensor modes to map its next low bit to the current bit of the linear index ($b^{t}$). 
Hence, the action space $A = \{a_{t} \in \mathbb{N}: 1 \leq a_{t} \leq N\}$ is discrete and limited to $N$ actions, one per tensor mode.
By restricting the number of permitted actions, \RELATE decreases the estimation error in selecting the highest-valued action~\cite{van2016deep}.
Importantly, by always mapping the next low bit of a mode index and preserving the internal bit order within each mode index, \RELATE dramatically reduces the space of possible linearized representations. 
Specifically, it transforms the problem of exploring all $\ell(p)!$ linearized formats to interleaving $N$ bit sequences, with length $\ell(n) ~\forall n \in \{1, 2, \dots, N\}$, into a sequence of length $\ell(p)$, which is the classical multinomial problem~\cite{stanley2011enumerative}. Since the bits of every mode index $i_{n}$ represent a subsequence of the linear encoding $p$, the number of ways to generate this interleaved linear encoding is given by:
\begin{equation}
\binom{\ell(p)}{\ell(1),\,\ell(2),\,\dots,\,\ell(N)}
   \;=\;
   \frac{\ell(p)!}{\ell(1)!\,\ell(2)!\,\dotsm \ell(N)!}
\end{equation}

Thus, for the toy example in Figure~\ref{fig:env}, \RELATE limits the number of actions to $3$ and reduces the state-action space from $720$ to $60$ linearized representations.   

\subsubsection{Transition Function}
Once an action $a_t$ is selected, the transition function $T(s_t, a_t, s_{t+1})$ constructs a new environment encoding $s_{t+1}$ with $t+1$ hot bits. If $s_{t+1}$ is a terminal state with $\ell(p)$ hot bits, the transition function invokes the reward function to examine the new linear encoding.     

\subsubsection{Reward Function}
To evaluate a newly constructed linear encoding, the reward function $R(s_t, a_t)$ sends it to the TD environment and receives execution time, as shown in Figure~\ref{fig:relate}. 
In this work, the TD environment extends the ALTO-based TD library to support custom linear encodings. During training, the environment benchmarks each candidate encoding across all tensor modes and reports the aggregate execution time.
However, using raw execution time as the reward leads to exploding/vanishing error gradients and unstable learning, even after extensive hyperparameter tuning. Rather than clipping the reward to a limited range~\cite{mnih2015human} and sacrificing fidelity, \RELATE retains magnitude information by evaluating the new linear encoding against an expert policy. Here, the expert policy is the ALTO-based encoding~\cite{laukemann2025accelerating}, and the reward function $R(s_t, a_t)$ generates a high-fidelity differential reward signal equal to the \emph{speedup} over that encoding, yielding bounded error gradients and allowing \RELATE to use the same hyperparameters across diverse sparse tensors.

\subsection{Learning Algorithm}
Although the proposed MDP formulation allows \RELATE to learn policies that minimize execution time, complex environments with many nonzero elements, extensive mode lengths, and high dimensionality remain challenging. The main reason is that the state-action space in such environments is still vast, even after substantial pruning, which in turn increases training cost. Moreover, reward evaluation is expensive, making it essential to minimize environment interactions and maximize the information extracted from every interaction. To address these challenges, \RELATE effectively combines model-free (double DQN~\cite{van2016deep}) and model-based DRL and introduces several mechanisms and optimizations to accelerate training. 

\begin{figure}[tb]
\centering
\includegraphics[width=1.0\linewidth]{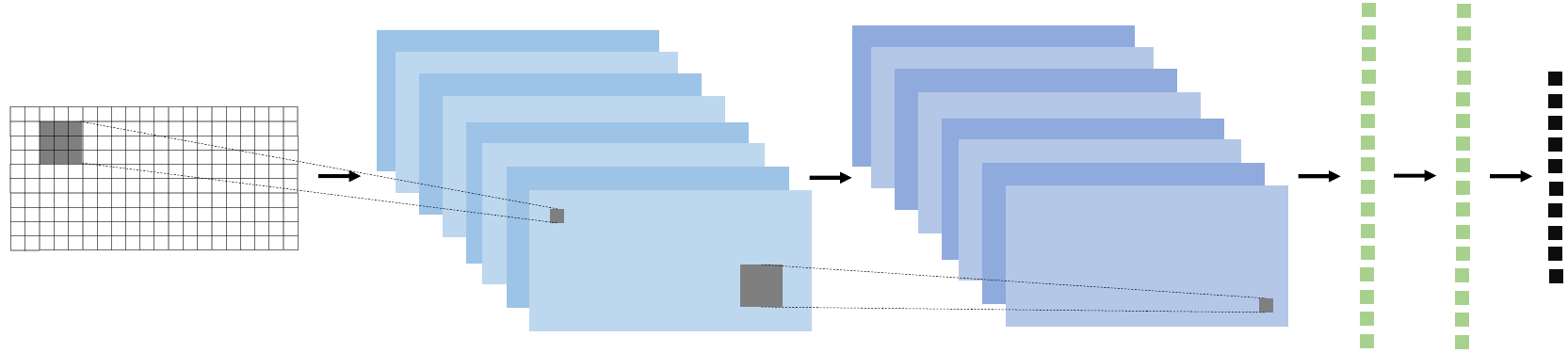}
	\caption{The proposed CNN-based policy/target network architecture, which adapts to the size of the state-action space. Specifically, in this architecture, the number of hidden units is proportional to the number of dimensions $N$ and the number of encoding bits $\ell(p)$. The input is the current $N\times\ell(p)$ encoding matrix, and the output contains one action value per tensor mode. Two convolutional layers with $16$ and $32$ feature maps and $3\times3$ kernels, followed by two fully connected layers, extract hierarchical bit-placement features for action-value prediction.} 
\label{fig:net}
\end{figure}

Algorithm~\ref{fig:relate-algo}, together with Figure~\ref{fig:relate}, outlines the key concepts of the proposed learning strategy. 
Given a sparse tensor \TENSOR{X} and a TD workload, \RELATE constructs an environment representation, based on the MDP formulation detailed in $\S$\ref{sec:approach-prob}, to find an optimized linear encoding $p^*$. \RELATE leverages a prioritized replay buffer~\cite{schaul2015prioritized} to improve sample efficiency by keeping a diverse set of significant recent transitions.
To learn a stable encoding policy, \RELATE uses a policy network ($\theta$) along with a target network ($\theta^-$) to learn a mapping from the high-dimensional state space $S$ to the discrete action space $A$. As illustrated in Figure~\ref{fig:net}, the proposed DRL agent uses an adaptive CNN architecture, in which the number of hidden units increases with the size of the state-action space without requiring per-tensor retuning. We chose this architecture because the environment state (encoding matrix) has a hierarchical spatial structure (sub-tensors) that can be exploited by the CNN to learn bit-placement patterns across intermediate states. Both the policy and target networks comprise two convolutional layers, with $16$ and $32$ feature maps respectively, each using $3\times3$ filters to capture the spatial correlations present within a receptive field of $5\times5$, followed by two fully connected layers. 
The input layer ingests an $N \times \ell(p)$ encoding matrix representing the current environment state, while the output layer emits $N$ activations, each indicating the value of a possible action. 

\begin{algorithm}[tb]
\begin{algorithmic}[1]
\footnotesize 
\Require An environment $\mathbf{ENV}$(\TENSOR{X}) for the sparse tensor \TENSOR{X} $\in$ \REALTHREE{I_{1}}{\dots}{I_{N}}
\Ensure Optimized linear encoding $p^*$
\State Initialize policy network with random weights $\theta$
\State Initialize target network weights: $\theta^- \leftarrow \theta$
\State Initialize prioritized experience replay buffer $\mathcal{B}^\mathrm{PER}$
\State Initialize elastic patience window $w \leftarrow E_{max}$
\For{episode $=1$ to $E_{max}$}
    \State Initialize episodic memory buffer $\mathcal{B}^\mathcal{E}$
    \State $s_t \leftarrow s_{0}$ \Comment{No encoding state}
    \For{$t = 0$ to $\ell(p)-1$}
    \State $A^- = \mathbb{GET\_VALID\_ACTIONS}(s_t)$ \label{algo:masking}
    \State With probability $\epsilon$ select a random action $a_t \in A^-$, otherwise 
    \[
    a_t = \operatorname{argmax}_{a_t \in A^-} Q(s_t,a_t;\theta)
    \] \label{algo:selection}
    \State $(r_t, s_{t+1}) = \mathbb{FILTER\_EXE\_ACTION}(a_t, s_t,$ $\mathbf{ENV}$(\TENSOR{X})$)$ \label{algo:filter}
    \State Store transition $(s_t, a_t, r_t, s_{t+1})$ in $\mathcal{B}^\mathcal{E}$
    \State $s_t \leftarrow s_{t+1}$
\EndFor
\If{$|\mathcal{B}^{\mathrm{PER}}| \geq \ell(p)$}
    \State Sample a minibatch  $\mathcal{B} = \{(s_i,a_i,r_i,s_i')\}_{i=1}^{\ell(p)}$ from $\mathcal{B}^\mathrm{PER}$
    \State Update $\theta$ via an optimizer step at learning rate $\alpha$ on the loss
    \[
      \frac{1}{|\mathcal{B}|}\sum_{i=1}^{|\mathcal{B}|}
\Bigl[
r_i+\gamma\,Q\!\bigl(s_i',
  \operatorname{argmax}_{a'}Q(s_i',a';\theta);\theta^{-}\bigr)
      -Q(s_i,a_i;\theta)
\Bigr]^2
    \]
\EndIf    
      \If{episode $\bmod \tau_{\mathrm{update}} = 0$} \label{algo:target}
        \State Update target network: $\theta^- \leftarrow \theta$
      \EndIf    
    \State Decay $\epsilon$ and $\alpha$ \label{algo:decay}
    \State Reduce patience window $w$ as $\epsilon$ decays \label{algo:patience-decay}   
    \State $p^* \leftarrow \mathbb{UPDATE\_ENCODING}(s_{\ell(p)}, r_{\ell(p)-1})$ \Comment{Best encoding}      
\If{$p^*$ is not updated for $w$ episodes} \label{algo:patience} 
    \State \textbf{break} \Comment{Elastic early stopping}
\EndIf    
    \State $\forall r_t \in \mathcal{B}^\mathcal{E}: r_t \leftarrow \log (r_{\ell(p)-1}) \big/ \ell(p)$ \label{algo:shaping} \Comment{Reward shaping}
    \State Store episodic memory $\mathcal{B}^\mathcal{E}$ in $\mathcal{B}^\mathrm{PER}$
\EndFor
\Return $p^*$
\end{algorithmic}
\caption{Learning algorithm for constructing an optimized sparse tensor encoding. }
\label{fig:relate-algo}
\end{algorithm}

The training process unfolds over up to $E_{max}$ episodes, each consisting of $\ell(p)$ steps. For every episode, \RELATE moves the environment from its initial state $s_0$ to a terminal state $s_{\ell(p)}$ and gathers transitions in a temporary memory buffer for further processing. In each step $t$, \RELATE ensures that the number of bits selected from a mode and mapped to the linear encoding (i.e., the number of hot bits in any row $n$ of the encoding matrix, as illustrated in Figure~\ref{fig:env} and $\S$\ref{sec:approach-prob}) cannot exceed the number of mode bits $\ell(n) ~\forall ~n \in \{1, 2, \ldots, N\}$. Hence, it masks the activations of invalid actions (line~\ref{algo:selection}) by computing the set of valid actions $A^-$ based on the current environment state (line~\ref{algo:masking}). This masking is a hard constraint enforced by construction: the agent cannot select a mode once its bits are exhausted, so every episode ends with a valid encoding. This validity guarantee is therefore formal, whereas the bounded execution time of valid encodings is established empirically (\S\ref{sec:performance}).
Because the number of possible terminal states is large, effective navigation of the solution space is crucial. While exploring new actions leads to better solutions, it may affect convergence and learning stability. In contrast, exploiting current knowledge (embedded in the policy network $\theta$) accelerates convergence but may lead to suboptimal results. Thus, \RELATE performs heavy exploration early in training by selecting a random action with high probability $\epsilon$, then gradually reduces both $\epsilon$ and the learning rate $\alpha$ to avoid catastrophic divergence (line~\ref{algo:decay}). 

Once a valid action $a_t$ is selected, 
either randomly or based on its estimated value, 
\RELATE determines whether interaction with the environment is needed for reward evaluation (line~\ref{algo:filter}). Specifically, \RELATE assigns a default zero reward to intermediate (non-terminal) actions $a_t \forall t \neq  \ell(p) - 1$ and evaluates rewards only for terminal actions $a_{\ell(p) - 1}$. During exploration, \RELATE learns to judge its own actions by gradually forming a simple reward model of the environment (omitted from Algorithm~\ref{fig:relate-algo} for brevity).
The reward model is a lightweight predictor trained from observed real rewards. It flattens a terminal state into a vector and uses a fully connected network with two hidden layers and one scalar output to predict the terminal reward.  
Once enough reward samples have been collected, \RELATE uses this reward model for action filtering only when its mean relative prediction error on $10$ unseen samples falls below $10\%$, corresponding to the $90\%$ accuracy threshold in Table~\ref{tab:hyperparameters}.
The agent then learns from both real actions (reward evaluation) and imagined actions (reward model). When an imagined action is rated as high-value by the reward model, i.e., its estimated reward is no worse than the best observed reward by an amount equal to the model’s error margin, the agent evaluates its real reward and updates the model accordingly. Even in complex environments, a simple reward predictor is sufficient, as shown in Figure~\ref{fig:error}, because it predicts only terminal rewards rather than the full transition dynamics of the environment.
Additionally, by inspecting the reward cache before computing any terminal reward, \RELATE avoids evaluating already seen action sequences. 

After executing every action $a_t$, \RELATE stores the resulting transition in temporary episodic memory. Once the episode terminates, it trains the policy network ($\theta$) via a gradient-based optimizer on a minibatch sampled from the replay buffer. Policy updates begin only after the replay buffer contains at least one minibatch. During training, the value of an action selected by the policy network $\theta$ is evaluated with the target network $\theta^-$, which is updated at a slower rate, as shown in line~\ref{algo:target}, to provide more reliable estimates of future values~\cite{van2016deep}. 
At the end of every episode, \RELATE updates the best sparse encoding discovered so far using the reward of the newly constructed encoding.
Rather than exhausting a fixed episode budget, \RELATE uses elastic early stopping: training terminates when the best observed reward fails to improve for a patience window $w$ (line~\ref{algo:patience}), which shrinks as exploration decays (line~\ref{algo:patience-decay}).
Higher patience is needed early in training, when exploration is heavy and action-value estimates are less reliable, whereas lower patience suffices later as action-selection uncertainty decreases.
Because DRL optimization loss is not a reliable proxy for policy performance, our early-stopping criterion is based on reward improvement.

Since rewards are generated only for terminal actions/states, the agent can suffer from unstable and delayed learning as the feedback signal is both sparse and ambiguous, making it hard to determine which earlier actions mattered. To tackle this issue, \RELATE introduces a reward shaping mechanism (line~\ref{algo:shaping}) that allocates credit $r_t$ from the sparse reward $r_{\ell(p)-1}$ to all actions leading up to that reward. 
Because each linear encoding is a multinomial interleaving of mode bits, the contribution of any single action is strongly coupled to the other bit-placement decisions. In our experiments, discounted or position-weighted credit distribution schemes introduced ordering bias, whereas uniform credit distribution yielded the best results with minimal complexity, after converting speedup/slowdown factors into positive/negative rewards via a log-scale mapping. Finally, \RELATE moves the updated transitions from temporary episodic memory to the prioritized replay buffer, which is used during training in future episodes.

\section{Evaluation}\label{sec:results}
We evaluate \RELATE against state-of-the-art expert formats, covering both mode-agnostic and mode-specific representations. We characterize TD performance on the fifth generation of Intel Xeon Scalable processors (Emerald Rapids, EMR).

\subsection{Experimental Setup}
\subsubsection{Test Platform}
The evaluation was conducted on dual-socket Intel Xeon Platinum 8592+ processors based on the EMR micro-architecture. Each socket has 64 physical cores, running at a fixed 1.9\,\GHZ ~base frequency, and 320\,\MiB ~L3 cache. The experiments use all ($128$) physical cores in the target system, with one thread pinned to each core and turbo mode disabled to ensure stable measurements.
The test system has 1024 GiB of DDR5 main memory and runs the Rocky Linux~10.0 distribution. The code is compiled with the Intel C/C++ compiler (v2025.3.0) using the optimization flags \texttt{-O3} 
\texttt{-qopt-zmm-usage=high}
\texttt{-xHost} to fully utilize vector units. The \RELATE agent is implemented in Python 3.12 and PyTorch 2.10, and communicates with the TD environment through mpi4py (v4.1.1) over Intel MPI (v2021.17). The \RELATE agent and TD environment run as separate MPI ranks on distinct nodes.

\subsubsection{Datasets}
The experiments use all real-world sparse tensors from FROSTT~\cite{frostt} whose working sets exceed cache capacity. For smaller tensors, in terms of dimensions and nonzero counts, the choice of linearized sparse encoding has negligible performance impact, and \RELATE matches the performance of the state-of-the-art ALTO format~\cite{laukemann2025accelerating}. Table~\ref{tab:problems} lists the target sparse tensors, sorted by number of nonzero elements (\#NNZs).     

\begin{table}[tb]
\centering
\caption{Characteristics of the sparse tensor data sets used in the evaluation, ordered by increasing number of nonzeros (\#NNZs). Dimensions list the length of each tensor mode.}
\label{tab:problems}
\begin{tabular}%
{|p{1.31cm}|p{3.685cm}|p{0.795cm}|p{1.395cm}|}
\hline
\textbf{Tensor} & \textbf{Dimensions} & \textbf{\#NNZs} & \textbf{Density} \\\toprule\hline
\textsc{darpa} & $22.5\mathrm{K}\times22.5\mathrm{K}\times23.8\mathrm{M}$& $28.4\mathrm{M}$& $2.4\times10^{-09}$ \\\hline 
\textsc{fb-m} & $23.3\mathrm{M}\times23.3\mathrm{M}\times166$ & $99.6\mathrm{M}$& $1.1\times10^{-09}$\\\hline 
\textsc{flickr-4d} & $319.7\mathrm{K}\times28.2\mathrm{M}\times1.6\mathrm{M}\times731$& $112.9\mathrm{M}$& $1.1\times10^{-14}$\\\hline 
\textsc{flickr-3d} & $319.7\mathrm{K}\times28.2\mathrm{M}\times1.6\mathrm{M}$& $112.9\mathrm{M}$& $7.8\times10^{-12}$\\\hline 
\textsc{deli-4d} & $532.9\mathrm{K}\times17.3\mathrm{M}\times2.5\mathrm{M}\times1.4\mathrm{K}$& $140.1\mathrm{M}$& $4.3\times10^{-15}$\\\hline 
\textsc{deli-3d} & $532.9\mathrm{K}\times17.3\mathrm{M}\times2.5\mathrm{M}$& $140.1\mathrm{M}$& $6.1\times10^{-12}$\\\hline 
\textsc{nell-1} & $2.9\mathrm{M}\times2.1\mathrm{M}\times25.5\mathrm{M}$& $143.6\mathrm{M}$& $9.1\times10^{-13}$\\\hline 
\textsc{amazon} & $4.8\mathrm{M}\times1.8\mathrm{M}\times1.8\mathrm{M}$& $1.7\mathrm{B}$ & $1.1\times10^{-10}$\\\hline 
\textsc{patents} & $46\times239.2\mathrm{K}\times239.2\mathrm{K}$& $3.6\mathrm{B}$ & $1.4\times10^{-03}$\\\hline 
\textsc{reddit} & $8.2\mathrm{M}\times177\mathrm{K}\times8.1\mathrm{M}$& $4.7\mathrm{B}$& $4.0\times10^{-10}$\\\hline 
\end{tabular}
\end{table}

\subsubsection{Configurations}
We compare \RELATE\footnote{Available at: 
\url{https://github.com/IntelLabs/ReLATE-Sparse}} to the best mode-agnostic and mode-specific sparse tensor representations, namely linearized~\cite{alto_2021, laukemann2025accelerating} and CSF-based~\cite{Smith2015, smith2017accelerating} formats, respectively. We use the state-of-the-art TD libraries: ALTO\footnote{Available at: 
\url{https://github.com/IntelLabs/ALTO}}
for mode-agnostic formats and SPLATT\footnote{Available at: \url{https://github.com/ShadenSmith/splatt}} for mode-specific formats, which have demonstrated superior performance~\cite{laukemann2025accelerating} to prior TD methods~\cite{kurt2022sparsity,li2018hicoo} for the target TD workloads. Additionally, ALTO serves as the basis for \RELATE’s TD environment and as the expert policy in its reward function, as detailed in $\S$\ref{sec:approach}. 
Because reward evaluation is expensive, we could not exhaustively tune \RELATE's hyperparameters using benchmark-based rewards. Instead, we performed this exploration in a sandbox mode that requires no real reward evaluation. Starting from an empty encoding, the agent is rewarded for reproducing the exact ALTO encoding in as few episodes as possible. Using this sandbox mode, we selected a single hyperparameter set (see Table~\ref{tab:hyperparameters}) applied across all tensors during real training, with no per-tensor tuning.
For best performance, SPLATT keeps $N$ tensor copies for an $N$-dimensional tensor, and we report the best results obtained across its two variants: standard CSF and CSF with tensor tiling. Likewise, ALTO selects the smallest linearized index size compatible with each target tensor (i.e., $64$ or $128$ bits).
In ALTO, we set the minimum fiber reuse for reduction-based synchronization to $8$ to improve performance on EMR. 
As in prior studies~\cite{Smith2015, alto_2021, laukemann2025accelerating}, we decompose each tensor data set into $16$ rank-one components using double-precision arithmetic and 64-bit integers. 
To benchmark TD performance, we measure the execution time of parallel MTTKRP over all tensor modes using $100$ iterations per measurement. We repeat each measurement $3$ times and report the median.
In the ``Shuffled'' experiments, we apply an independent, uniformly random permutation of the index labels (coordinates) in each mode, generated with a fixed seed, retaining the tensor's values, shape, and number of nonzero elements.

\begin{table}[tb]
\centering
\caption{\RELATE Hyperparameters.}
\label{tab:hyperparameters}
\begin{tabular}{|p{1.9cm}|p{0.7cm}|p{5.0cm}|}
\hline
\textbf{Hyperparameter} & \textbf{Value} & \textbf{Description}\\\toprule\hline
 $\gamma$    &   $0.99$       & Discount factor for future rewards. \\\hline 
 $|\mathcal{B}|$                    &     $\ell(p)$       & Number of samples per minibatch. \\\hline 
$|\mathcal{B}^\mathrm{PER}|$   &  $1\mathrm{M}$      & Maximum number of observations in the replay buffer. \\\hline 
Initial $\alpha$      &  $0.001$     & Initial learning rate before decay. \\\hline 
Min. $\alpha$      &  $0.0001$     & Learning rate after decay. \\\hline 
Initial $\epsilon$                &    $1.0$      & Initial value for $\epsilon$ before decay. \\\hline 
Min. $\epsilon$  &   $0.1$        & Value of $\epsilon$ after decay. \\\hline 
$\tau_{\mathrm{update}}$       & $100$ & Target update frequency. \\\hline 
$E_{max}$       & $5000$ & Maximum number of training episodes. \\\hline 
Min. accuracy              &   $90\%$   & Accuracy threshold for the predictive reward model. \\\hline 
Optimizer &   Adam   & Training optimizer used for model updates. \\\hline 
\end{tabular}
\end{table}

\subsection{Performance Results and Analysis}\label{sec:performance}

\begin{figure*}[tb]
\centering
\includegraphics[width=1.0\linewidth]{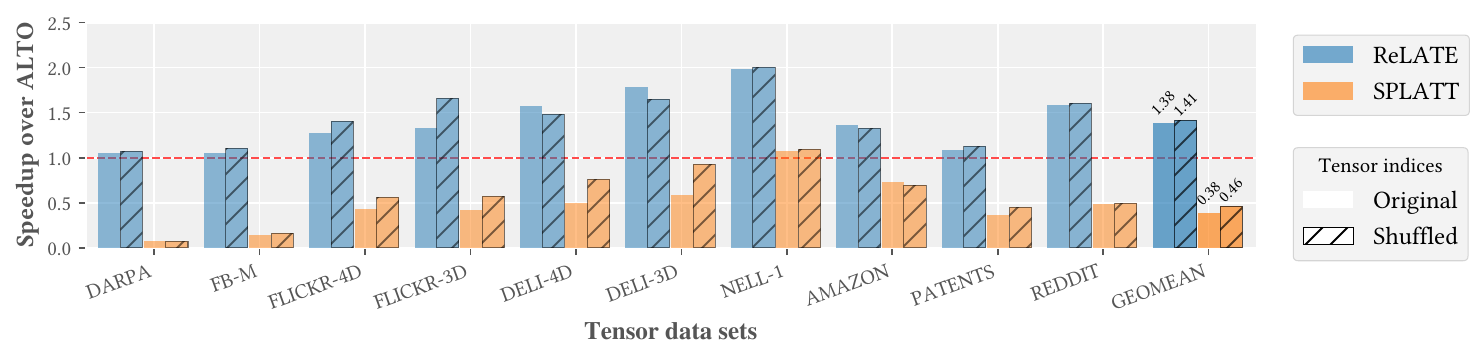}
	\caption{The performance of our DRL-based sparse encoding (\RELATE) and SPLATT relative to ALTO on a 128-core EMR system. The sparse tensors are sorted in increasing order of their size (number of nonzero elements). The right-most group of bars shows the geometric-mean speedup over ALTO.}
\label{fig:speedup}
\end{figure*}

\begin{figure*}[tb]
\centering
\includegraphics[width=1.0\linewidth]{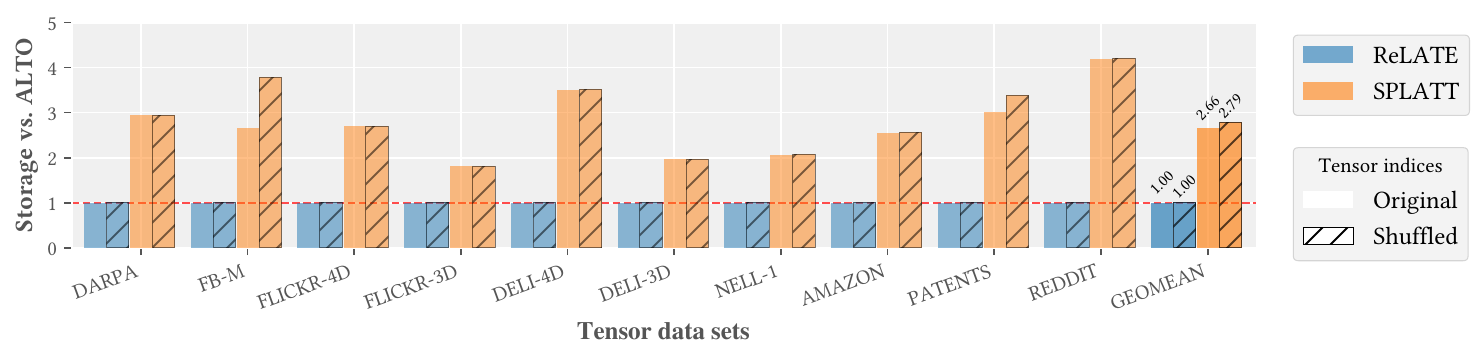}
	\caption{The required tensor storage across sparse tensor representations relative to the mode-agnostic ALTO format. The sparse tensors are sorted in increasing order of their size (number of nonzero elements). The right-most group of bars shows the geometric-mean storage ratio vs. ALTO.}
\label{fig:mem}
\end{figure*}

Figures~\ref{fig:speedup} and~\ref{fig:mem} compare the performance and storage of the sparse tensor representations found by \RELATE to prior state-of-the-art formats~\cite{laukemann2025accelerating}. Since expert-designed formats can be highly sensitive to the data distributions of sparse tensors~\cite{alto_2021}, we analyze the performance under random permutations of tensor indices (``Shuffled'') to break down inherent data skewness, in addition to the original tensors. The results show that \RELATE outperforms the best mode-agnostic and mode-specific formats across all sparse tensors, while matching the memory storage of the mode-agnostic ALTO format. Because the ALTO encoding is constructed by partitioning the tensor space along the longest mode first (Figure~\ref{fig:alto}), it corresponds to a greedy policy in our MDP (\S\ref{sec:approach-prob}). Hence, the reported speedups over ALTO quantify the benefit of our learned policy over greedy format construction.

Specifically, \RELATE realizes a geometric-mean speedup of $1.38$--$1.41\times$, and up to $2\times$ speedup, over the best expert-designed format. Compared to SPLATT, \RELATE achieves up to $14.5\times$ speedup, with a geometric-mean speedup of $3.05$--$3.56\times$.    
These performance gains hold for both the original and randomly permuted sparse tensors, reflecting the adaptive nature of the learned encoding.
SPLATT has better average performance on randomly permuted tensors compared to the original ones, as random permutation reduces data skewness and improves the workload balance of CSF-based formats; however, both ALTO and \RELATE significantly outperform SPLATT because of the workload balance, high data reuse, and low synchronization cost of linearized formats~\cite{alto_2021, laukemann2025accelerating}. 

Additionally, Figure~\ref{fig:speedup} and Table~\ref{tab:problems} show that the performance gains of our learned sparse encoding over the best expert format increase with the tensor size and sparsity. This performance trend not only highlights the limitations of prior input-oblivious heuristics for constructing sparse tensor formats, but also indicates that \emph{learning-augmented algorithms are crucial for tackling low-density, large-scale tensors}, which are particularly challenging to optimize on modern parallel processors, due to their limited data reuse, low arithmetic intensity, and random memory access~\cite{choi2018blocking, blco_2022, laukemann2025accelerating}.
Although \textsc{patents} is large, in terms of nonzero elements, its dimensions are short, so the performance gap between \RELATE and ALTO is small because factorization data is comparable in size to the L$3$ cache. While \textsc{deli-3d} and \textsc{flickr-3d} have higher density than \textsc{deli-4d} and \textsc{flickr-4d}, \RELATE performs even better than the expert formats in the former cases compared to
the latter. Further analysis reveals that the density increase stems from removing the $4^{\mathrm{th}}$ dimension in these tensors, which represents dates and is extremely short. Because TD operations along this mode exhibit high fiber reuse, removing it exacerbates the challenges of executing TD operations along the remaining longer modes, widening the performance gap between \RELATE and the prior input-oblivious formats.           

Because the mode-agnostic ALTO uses a single tensor copy to perform TD operations across all modes, its memory storage is significantly lower than the mode-specific SPLATT, which keeps one tensor copy per mode to maximize performance. Since the storage of linearized formats depends on tensor shape rather than data distribution, unlike SPLATT and other compressed formats~\cite{laukemann2025accelerating}, \RELATE matches the storage requirements of ALTO for the original and randomly permuted tensor data sets. 
Figure~\ref{fig:mem} shows that SPLATT consumes up to $4.2\times$ more storage than \RELATE and ALTO. Although random permutation of tensor indices improves workload balance and performance for SPLATT, it also increases its average storage requirement relative to the original tensors.      

\subsubsection{Roofline Analysis}
To gain insight into \RELATE's performance relative to the prior state-of-the-art format, we constructed a Roofline model~\cite{roofline:2009} for the target EMR processors and collected performance counters with LIKWID~v5.4.1~\cite{likwid} across representative data sets. 
Under the Roofline model, performance is bounded by \(\Phi = \mathrm{min}(\Phi_{\mathrm{peak}}, BW_{\mathrm{peak}} \times OI)\), with
\(\Phi_{\mathrm{peak}}\) denoting the theoretical peak compute throughput, \(BW_{\mathrm{peak}}\) peak memory bandwidth, and \(OI\) operational intensity (FLOPs per byte).
We estimate \(\Phi_{\mathrm{peak}}\) 
assuming a steady state, in which a core can issue two fused multiply-add~(FMA) instructions per cycle on 512-bit 
(eight double-precision) 
vector registers, delivering up to 32 double-precision FLOPs per cycle per core.
We augment our Roofline model to account for the cache bandwidth, as in prior work~\cite{laukemann2025accelerating}. We measure L$2$/L$3$ and main-memory bandwidth with \texttt{likwid-bench} from the LIKWID tool suite, and adopt the theoretical L1 bandwidth: two cache lines (512-bit transactions) per cycle per core. 

\begin{figure}[tb]
    \centering
\includegraphics[width=0.98\linewidth]{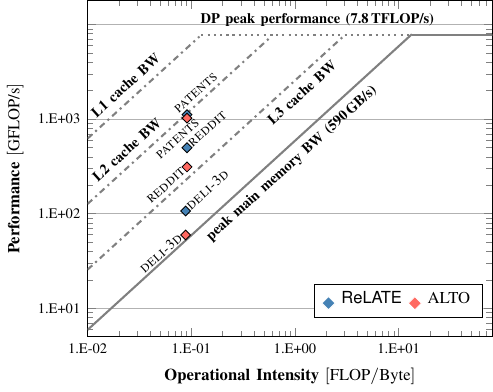}    
    \caption{Cache-aware Roofline analysis of TD operations using \RELATE (blue diamonds) and ALTO (red diamonds) on a 128-core EMR system. Higher vertical position indicates greater floating-point throughput.} 
    \label{fig:roofline}
\end{figure}

The learned linearized encoding does not change the synchronization mechanism used by TD operations like MTTKRP, which is determined by tensor characteristics (shape and mode lengths) rather than the encoding~\cite{laukemann2025accelerating}. Thus, \RELATE's gains come from better cache utilization, lower memory traffic, and higher floating-point throughput.
Figure~\ref{fig:roofline} and Table~\ref{tab:mem} detail the performance of three representative tensors. Because \textsc{deli-3d} is highly sparse and has limited data reuse, its performance using ALTO is largely bounded by main-memory bandwidth. Nevertheless, \RELATE discovers a sparse encoding that significantly decreases the L$3$ miss ratio, reducing memory volume by $43\%$. Although \textsc{reddit} is a low-density tensor, its larger number of nonzero elements relative to its dimensions increases data reuse compared to \textsc{deli-3d}. As a result, the performance of \textsc{reddit} using ALTO is bounded by L$3$ cache bandwidth. Even in this case, \RELATE improves the L$3$ hit ratio and reduces memory traffic by $39\%$. In contrast to the previous cases, \textsc{patents} has higher density because of its small dimensions, so its factorization data is served largely from L$2$ cache, as reflected by its much lower memory volume than \textsc{reddit} even though both tensors have a comparable number of nonzero elements. Thus, while \RELATE slightly reduces the L$2$ miss ratio, the performance gain is limited compared to the other large-scale, low-density tensors.             

\begin{table}[tb]
    \centering
    \caption{Comparison of the memory-related metrics (obtained by perf counters) across \RELATE and ALTO on a 128-core EMR system for three representative tensors.}
\begin{tabular}{|p{1.2cm}|p{1.2cm}|p{1.5cm}|p{1.1cm}|p{1.1cm}|}
\hline
\textbf{Tensor} & \textbf{Format} & \textbf{Memory Volume} & \textbf{L$2$ Miss Ratio} & \textbf{L$3$ Miss Ratio}\\\toprule\hline
\multirow{2}{*}{{\textsc{deli-3d}}} & 
\RELATE & $52.88~\text{GB}$ & $69.42\%$  & $55.64\%$\\ & 
ALTO & $93.43~\text{GB}$ & $75.21\%$  & $78.63\%$
\\\hline
\multirow{2}{*}{{\textsc{patents}}} &   
\RELATE   & $177.23~\text{GB}$ & $53.46\%$  & $46.83\%$  \\ & 
ALTO &  $175.98~\text{GB}$ & $55.80\%$  & $58.30\%$
\\\hline
\multirow{2}{*}{{\textsc{reddit}}} &   
\RELATE   & $ 438.18~\text{GB}$  & $43.84\%$  & $ 33.90\%$ \\ &
ALTO&  $716.01~\text{GB}$ & $48.47\%$  & $42.58\%$ 
\\\hline         
\end{tabular}
    \label{tab:mem}
\end{table}

\subsection{Agent Effectiveness}

\begin{figure}[tb]
\centering
\includegraphics[width=1.0\linewidth]{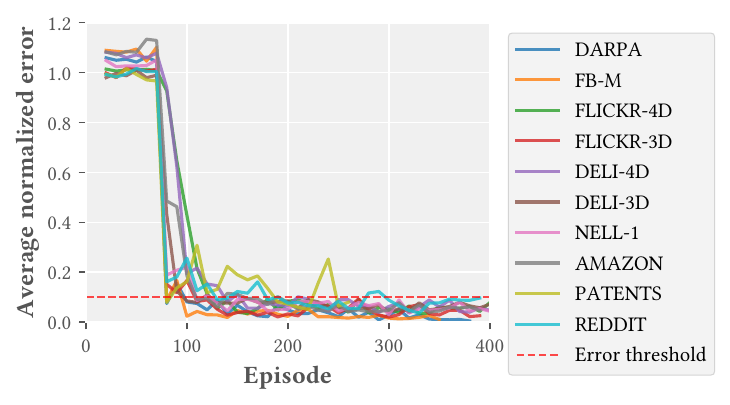}
	\caption{Average estimation error of \RELATE's reward model, relative to the actual reward, over the first $400$ episodes. 
    Error is measured on $10$ held-out test samples.
    The $10\%$ threshold corresponds to the required $90\%$ model accuracy. The reward model reaches this threshold within $300$ episodes. Error curves can end before the episode limit due to elastic early stopping.}
\label{fig:error}
\end{figure}

To evaluate the effectiveness of the \RELATE agent, we validate its reward model and categorize its actions accordingly.  
Figure~\ref{fig:error} shows the average estimation error of the reward model across the first $400$ training episodes, measured as the absolute difference between the estimated and actual reward and normalized by the magnitude of the actual reward. 
The reported error is averaged over $10$ test samples not seen during training. 
The results show that the reward model quickly learns the environment dynamics, 
with the estimation error falling below the threshold within $300$ episodes.
Since training employs elastic span (early stopping) with patience proportional to the exploration rate, error trajectories end at different episodes based on each tensor’s convergence.

\begin{figure}[tb]
\centering
\includegraphics[width=1.0\linewidth]{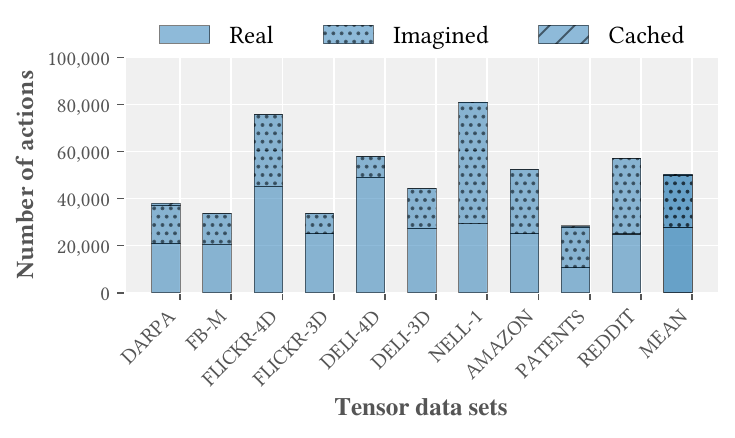}
	\caption{Number of real, cached, and imagined actions taken by the \RELATE agent across sparse tensors, sorted by increasing number of nonzero elements.
    Real actions invoke the TD environment to measure rewards, cached actions reuse previously measured rewards, and imagined actions use rewards estimated by the reward model. Thus, the cached and imagined categories quantify the benchmark evaluations avoided by \RELATE.
    The right-most group of bars reports the mean number of actions in each category.}
\label{fig:actions}
\end{figure}

Figure~\ref{fig:actions} shows the number and type of actions taken by the agent across sparse tensor environments. Actions are categorized into real, imagined, or cached, and only real actions require reward evaluation.
The results show that real actions represent $55.6\%$ of all actions on average, while the remaining $44.4\%$ are resolved by the reward model or cache without benchmarking, reflecting the effectiveness of our action filtering mechanisms. Model-based filtering, in which the agent uses imagined/estimated reward, is crucial in most environments, with imagined actions forming the majority on the largest tensors (\textsc{nell-1}, \textsc{amazon}, \textsc{patents}, and \textsc{reddit}), while cache-based filtering is primarily useful in simple environments, like \textsc{darpa}.          

\subsection{Training Overhead}
\begin{figure}[tb]
    \centering
\includegraphics[width=1.0\linewidth]{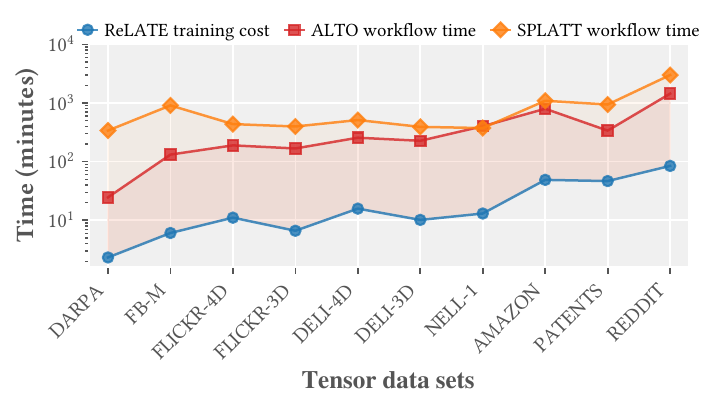}
    \caption{Offline training cost of \RELATE compared with a representative TD workflow time using ALTO and SPLATT on a 128-core EMR system. Times are reported in minutes on a logarithmic scale. When calculated as a geometric mean across all tensors, the one-time training cost is merely $5.82\%$ of the ALTO workflow time and $2.3\%$ of the SPLATT workflow time.}
    \label{fig:overhead}
\end{figure}

Figure~\ref{fig:overhead} compares the offline training cost of \RELATE to the time required to execute a representative TD workflow using ALTO and SPLATT on the target system. In practice, a typical TD workflow executes TD with several randomly initialized factor matrices, spanning a wide range of principal components/factors, and repeats this process with different data-processing methods to overcome noise, outliers, non-linear artifacts, and systematic biases~\cite{10.1145/3519383}. 
Even for relatively small data sets, this can require $400$ TD runs per data-processing method~\cite{10.1145/3519383}, each taking hundreds of iterations to reach convergence~\cite{KoBa09, Smith2015, 10.1145/3519383, 10.1002/nla.2511}, as noted in~$\S$\ref{sec:background-TD}. Conservatively assuming $100$ 
iterations per run, a single data-processing method amounts to $40{,}000$ iterations, which is the workflow baseline used in Figure~\ref{fig:overhead}. For real-world tensors with multiple data-processing methods, the total number of MTTKRP operations is in the hundreds of thousands.

The results show that \RELATE achieves low offline training overhead, due to our effective action masking, action filtering, and elastic training. Specifically, the geometric-mean training overhead is only $5.82\%$ of the corresponding ALTO workflow time and $2.3\%$ of the SPLATT workflow time.
The higher relative
overhead with respect to ALTO reflects its significantly faster per-iteration execution
compared to SPLATT. 
Moreover, this comparison is conservative: \RELATE's training time includes sequential steps such as format generation, whereas the ALTO and SPLATT workflow times account only for computation, excluding tensor reading, preprocessing, and format construction. The sequential training steps can be further optimized, and multiple server nodes can evaluate rewards concurrently, reducing training time beyond the single-server setup used here.
Because offline training is a one-time cost, the learned encoding is deployed with no additional overhead. This one-time cost may be further amortized across related tensors, since those produced by similar generating processes are likely to share an effective linearization strategy. However, systematically evaluating such an amortization opportunity is difficult given the scarcity of public large-scale tensors.

\section{Conclusion}
This work introduces a novel approach for optimizing sparse tensor decomposition through reinforcement learning techniques, eliminating the need for labeled training samples. By effectively learning high-quality sparse tensor representations, our \RELATE framework delivers substantial performance gains across diverse data sets, achieving up to $2\times$ speedup over the best expert-designed formats. 
\RELATE is trained through direct interaction with the target environment, with the agent and environment placed on separate compute nodes to isolate reward measurements from learning activities. Across tensors, its one-time training cost has a geometric mean of only 5.82\% relative to the workflow time, and it always generates functionally correct sparse encodings.
The learned sparse representations address the key performance bottlenecks, yielding better cache utilization, lower memory traffic, and higher floating-point throughput, especially for large-scale, low-density tensors that are hard to optimize on modern parallel processors.
Because the reward is measured on the target machine, \RELATE adapts to different cache, memory, and core-count configurations without changing its problem formulation, so porting to another architecture requires only one-time offline retraining. Future work will investigate transferring these learning capabilities to related sparse tensor operations and distributed-memory architectures. 

\balance

\section*{Acknowledgment}
ChatGPT and Claude were used to proofread the text and assist with code development. No AI-generated text has been used verbatim. All code was designed, developed, and validated by the authors.

\bibliographystyle{IEEEtran}
\bibliography{relate}

@inproceedings{tuncer2017diagnosing,
  title={{Diagnosing Performance Variations in HPC Applications using Machine Learning}},
  author={Tuncer, Ozan and Ates, Emre and Zhang, Yijia and Turk, Ata and Brandt, Jim and Leung, Vitus J and Egele, Manuel and Coskun, Ayse K},
  booktitle={High Performance Computing: 32nd International Conference, ISC High Performance 2017, Frankfurt, Germany, June 18--22, 2017, Proceedings 32},
  pages={355--373},
  year={2017},
  address = {Switzerland},
  publisher = {Springer},
  note={{DOI}: \href{https://doi.org/10.1007/978-3-319-58667-0_19}{10.1007/978-3-319-58667-0\_19}},
}

@ARTICLE{laukemann2025accelerating,
  author={Laukemann, Jan and Helal, Ahmed E. and Anderson, S. Isaac Geronimo and Checconi, Fabio and Soh, Yongseok and Tithi, Jesmin Jahan and Ranadive, Teresa and Gravelle, Brian J. and Petrini, Fabrizio and Choi, Jee},
  journal={IEEE Transactions on Parallel and Distributed Systems}, 
  title={{Accelerating Sparse Tensor Decomposition Using Adaptive Linearized Representation}}, 
  year={2025},
  volume={36},
  number={5},
  pages={1025--1041},
  note={{DOI}: \href{https://doi.org/10.1109/TPDS.2025.3553092}{10.1109/TPDS.2025.3553092}}
}

@inproceedings{hessel2018rainbow,
author = {Hessel, Matteo and Modayil, Joseph and van Hasselt, Hado and Schaul, Tom and Ostrovski, Georg and Dabney, Will and Horgan, Dan and Piot, Bilal and Azar, Mohammad and Silver, David},
title = {{Rainbow: Combining Improvements in Deep Reinforcement Learning}},
year = {2018},
isbn = {978-1-57735-800-8},
publisher = {AAAI Press},
booktitle = {Proceedings of the Thirty-Second AAAI Conference on Artificial Intelligence and Thirtieth Innovative Applications of Artificial Intelligence Conference and Eighth AAAI Symposium on Educational Advances in Artificial Intelligence},
articleno = {393},
numpages = {8},
pages={3215--3222},
location = {New Orleans, Louisiana, USA},
address = {USA},
series = {AAAI'18/IAAI'18/EAAI'18},
note={{DOI}: \href{https://doi.org/10.1609/aaai.v32i1.11796}{10.1609/aaai.v32i1.11796}}
}

@misc{schaul2015prioritized,
      title={{Prioritized Experience Replay}}, 
      author={Tom Schaul and John Quan and Ioannis Antonoglou and David Silver},
      year={2016},
      eprint={1511.05952},
      archivePrefix={arXiv},
      primaryClass={cs.LG},
      note={{DOI}: \href{https://doi.org/10.48550/arXiv.1511.05952}{10.48550/arXiv.1511.05952}},      
}

@inproceedings{van2016deep,
author = {van Hasselt, Hado and Guez, Arthur and Silver, David},
title = {{Deep Reinforcement Learning with Double Q-Learning}},
year = {2016},
publisher = {AAAI Press},
booktitle = {Proceedings of the Thirtieth AAAI Conference on Artificial Intelligence},
pages = {2094--2100},
numpages = {7},
location = {Phoenix, Arizona},
address = {USA},
series = {AAAI'16},
note={{DOI}: \href{https://doi.org/10.1609/aaai.v30i1.10295}{10.1609/aaai.v30i1.10295}}
}

@article{mnih2015human,
  title={{Human-Level Control through Deep Reinforcement Learning}},
  author={Mnih, Volodymyr and Kavukcuoglu, Koray and Silver, David and Rusu, Andrei A and Veness, Joel and Bellemare, Marc G and Graves, Alex and Riedmiller, Martin and Fidjeland, Andreas K and Ostrovski, Georg and others},
  journal={Nature},
  volume={518},
  number={7540},
  pages={529--533},
  year={2015},
  publisher={Nature Publishing Group},
  note={{DOI}: \href{https://doi.org/10.1038/nature14236}{10.1038/nature14236}},
}

@misc{mnih2013playing,
      title={{Playing Atari with Deep Reinforcement Learning}}, 
      author={Volodymyr Mnih and Koray Kavukcuoglu and David Silver and Alex Graves and Ioannis Antonoglou and Daan Wierstra and Martin Riedmiller},
      year={2013},
      eprint={1312.5602},
      archivePrefix={arXiv},
      primaryClass={cs.LG},
      note={{DOI}: \href{https://doi.org/10.48550/arXiv.1312.5602}{10.48550/arXiv.1312.5602}},
}

@book{bennett2023brief,
  title={{A Brief History of Intelligence: Evolution, AI, and The Five Breakthroughs That Made Our Brains}},
  author={Bennett, Max S},
  year={2023},
  isbn={9780063286344},
  lccn={2023008730},
  address = {USA},
  publisher={HarperCollins}
}

@article{mitzenmacher2022algorithms,
  title={{Algorithms with Predictions}},
  author={Mitzenmacher, Michael and Vassilvitskii, Sergei},
  journal={Communications of the ACM},
  volume={65},
  number={7},
  pages={33--35},
  year={2022},
  publisher={ACM New York, NY, USA},
  note={{DOI}: \href{https://doi.org/10.1145/3528087}{10.1145/3528087}},
}

@article{panagakis2021tensor,
  title={{Tensor Methods in Computer Vision and Deep Learning}},
  author={Panagakis, Yannis and Kossaifi, Jean and Chrysos, Grigorios G and Oldfield, James and Nicolaou, Mihalis A and Anandkumar, Anima and Zafeiriou, Stefanos},
  journal={Proceedings of the IEEE},
  volume={109},
  number={5},
  pages={863--890},
  year={2021},
  publisher={IEEE},
  note={{DOI}: \href{https://doi.org/10.1109/JPROC.2021.3074329}{10.1109/JPROC.2021.3074329}},
}

@article{sun2021input,
  title={{Input-Aware Sparse Tensor Storage Format Selection for Optimizing MTTKRP}},
  author={Sun, Qingxiao and Liu, Yi and Yang, Hailong and Dun, Ming and Luan, Zhongzhi and Gan, Lin and Yang, Guangwen and Qian, Depei},
  journal={IEEE Transactions on Computers},
  volume={71},
  number={8},
  pages={1968--1981},
  year={2021},
  publisher={IEEE},
  note={{DOI}: \href{https://doi.org/10.1109/TC.2021.3113028}{10.1109/TC.2021.3113028}},
}

@article{yadav2018mining,
  title={{Mining Electronic Health Records (EHRs): A Survey}},
  author={Yadav, Pranjul and Steinbach, Michael and Kumar, Vipin and Simon, Gyorgy},
  journal={ACM Computing Surveys (CSUR)},
  volume={50},
  number={6},
  pages={1--40},
  year={2018},
  publisher={ACM New York, NY, USA},
  note={{DOI}: \href{https://doi.org/10.1145/3127881}{10.1145/3127881}},
}

@inproceedings{alto_2021,
    author = {Helal, Ahmed E. and Laukemann, Jan and Checconi, Fabio and Tithi, Jesmin Jahan and Ranadive, Teresa and Petrini, Fabrizio and Choi, Jeewhan},
    title = {{ALTO: Adaptive Linearized Storage of Sparse Tensors}},
    year = {2021},
    isbn = {9781450383356},
    publisher = {Association for Computing Machinery},
    address = {New York, NY, USA},
    note = {{DOI}: \href{https://doi.org/10.1145/3447818.3461703}{10.1145/3447818.3461703}},
    booktitle = {Proceedings of the 35th ACM International Conference on Supercomputing},
    pages = {404--416},
    numpages = {13},
    location = {Virtual Event, USA},
    series = {ICS '21}
}

@inproceedings{blco_2022,
    author = {Nguyen, Andy and Helal, Ahmed E. and Checconi, Fabio and Laukemann, Jan and Tithi, Jesmin Jahan and Soh, Yongseok and Ranadive, Teresa and Petrini, Fabrizio and Choi, Jee W.},
    title = {{Efficient, Out-of-Memory Sparse MTTKRP on Massively Parallel Architectures}},
    year = {2022},
    isbn = {9781450392815},
    publisher = {Association for Computing Machinery},
    address = {New York, NY, USA},
    note = {{DOI}: \href{https://doi.org/10.1145/3524059.3532363}{10.1145/3524059.3532363}},
    booktitle = {Proceedings of the 36th ACM International Conference on Supercomputing},
    articleno = {26},
    pages={1--13},
    numpages = {13},
    location = {Virtual Event},
    series = {ICS '22}
}

@inproceedings{symeonidis2016matrix,
    title={{Matrix and Tensor Decomposition in Recommender Systems}},
    author={Symeonidis, Panagiotis},
    booktitle={Proceedings of the 10th ACM Conference on Recommender Systems},
    pages={429--430},
    year={2016},
    publisher = {Association for Computing Machinery},
    address = {New York, NY, USA},
    note={{DOI}: \href{https://doi.org/10.1145/2959100.2959195}{10.1145/2959100.2959195}},
}

@ARTICLE{sidiropoulos2017tensor,
    author={Sidiropoulos, Nicholas D. and De Lathauwer, Lieven and Fu, Xiao and Huang, Kejun and Papalexakis, Evangelos E. and Faloutsos, Christos},
    journal={IEEE Transactions on Signal Processing}, 
    title={{Tensor Decomposition for Signal Processing and Machine Learning}}, 
    year={2017},
    volume={65},
    number={13},
    pages={3551--3582},
    note = {{DOI}: \href{https://doi.org/10.1109/TSP.2017.2690524}{10.1109/TSP.2017.2690524}}
}

@article{kobayashi2018extracting,
    title={{Extracting The Multi-Timescale Activity Patterns of Online Financial Markets}},
    author={Kobayashi, Teruyoshi and Sapienza, Anna and Ferrara, Emilio},
    journal={Scientific Reports},
    volume={8},
    number={1},
    pages = {11184},
    year={2018},
    publisher={Nature Publishing Group},
    note = {{DOI}: \href{https://doi.org/10.1038/s41598-018-29537-w}{10.1038/s41598-018-29537-w}},
}

@inproceedings{li2018hicoo,
    author={Li, Jiajia and Sun, Jimeng and Vuduc, Richard},
    booktitle={{SC18: International Conference for High Performance Computing, Networking, Storage and Analysis}}, 
    title={{HiCOO: Hierarchical Storage of Sparse Tensors}}, 
    year={2018},
    volume={},
    number={},
    pages={238--252},
    address = {USA},
    publisher={IEEE},
    note={{DOI}: \href{https://doi.org/10.1109/SC.2018.00022}{10.1109/SC.2018.00022}},
}

@inproceedings{li2019efficient,
    author = {Li, Jiajia and U\c{c}ar, Bora and \c{C}ataly\"{u}rek, \"{U}mit V. and Sun, Jimeng and Barker, Kevin and Vuduc, Richard},
    title = {{Efficient and Effective Sparse Tensor Reordering}},
    year = {2019},
    isbn = {9781450360791},
    publisher = {Association for Computing Machinery},
    address = {New York, NY, USA},
    note = {{DOI}: \href{https://doi.org/10.1145/3330345.3330366}{10.1145/3330345.3330366}},
    booktitle = {Proceedings of the ACM International Conference on Supercomputing},
    pages = {227--237},
    numpages = {11},
    location = {Phoenix, Arizona},
    series = {ICS '19}
}

@inproceedings{nisa2019efficient,
    author = {Nisa, Israt and Li, Jiajia and Sukumaran-Rajam, Aravind and Rawat, Prasant Singh and Krishnamoorthy, Sriram and Sadayappan, P.},
    title = {{An Efficient Mixed-Mode Representation of Sparse Tensors}},
    year = {2019},
    isbn = {9781450362290},
    publisher = {Association for Computing Machinery},
    address = {New York, NY, USA},
    note = {{DOI}: \href{https://doi.org/10.1145/3295500.3356216}{10.1145/3295500.3356216}},
    booktitle = {Proceedings of the International Conference for High Performance Computing, Networking, Storage and Analysis},
    articleno = {49},
    pages={1--25},
    numpages = {25},
    location = {Denver, Colorado},
    series = {SC '19}
}

@INPROCEEDINGS{nisa2019load,
    author={Nisa, Israt and Li, Jiajia and Sukumaran-Rajam, Aravind and Vuduc, Richard and Sadayappan, P.},
    booktitle={2019 IEEE International Parallel and Distributed Processing Symposium (IPDPS)}, 
    title={{Load-Balanced Sparse MTTKRP on GPUs}}, 
    year={2019},
    volume={},
    number={},
    pages={123--133},
    address = {USA},
	publisher = {IEEE},
    note={{DOI}: \href{https://doi.org/10.1109/IPDPS.2019.00023}{10.1109/IPDPS.2019.00023}}
}

@INPROCEEDINGS{sun2020sptfs,
    author={Sun, Qingxiao and Liu, Yi and Dun, Ming and Yang, Hailong and Luan, Zhongzhi and Gan, Lin and Yang, Guangwen and Qian, Depei},
    booktitle={SC20: International Conference for High Performance Computing, Networking, Storage and Analysis}, 
    title={{SpTFS: Sparse Tensor Format Selection for MTTKRP via Deep Learning}}, 
    year={2020},
    volume={},
    number={},
    pages={1--14},
    address = {USA},
	publisher = {IEEE},
    note={{DOI}: \href{https://doi.org/10.1109/SC41405.2020.00022}{10.1109/SC41405.2020.00022}},
}

@misc{frostt,
    title = {{FROSTT}: The Formidable Repository of Open Sparse Tensors and Tools},
    author = {Smith, Shaden and Choi, Jee W. and Li, Jiajia and Vuduc, Richard and Park, Jongsoo and Liu, Xing and Karypis, George},
    url = {http://frostt.io/},
    year = {2017},
}

@article{peano1890courbe,
    title={{Sur une courbe, qui remplit toute une aire plane}},
    author={Peano, Giuseppe},
    journal={Mathematische Annalen},
    volume={36},
    number={1},
    pages={157--160},
    year={1890},
    month=mar,
    publisher={Springer},
    note={{DOI}: \href{https://doi.org/10.1007/BF01199438}{10.1007/BF01199438}}
}

@misc{likwid,
    author       = {Thomas Gruber and Michael Panzlaff and Jan Eitzinger and Georg Hager and Gerhard Wellein},
    title        = {{LIKWID}},
    month        = dec,
    year         = 2024,
    publisher    = {Zenodo},
    version      = {v5.4.1},
    howpublished = {Zenodo, version 5.4.1},  
    note         = {{DOI}: \href{https://zenodo.org/records/14364500}{10.5281/zenodo.14364500}}
}

@article{roofline:2009,
    author = {Williams, Samuel and Waterman, Andrew and Patterson, David},
    title = {{Roof{}line: An insightful Visual Performance Model for Multicore Architectures}},
    journal = {Commun. ACM},
    volume = {52},
    number = {4},
    year = {2009},
    issn = {0001-0782},
    pages = {65--76},
    publisher = {ACM},
    address = {New York, NY, USA},
    note = {{DOI}: \href{http://doi.org/10.1145/1498765.1498785}{10.1145/1498765.1498785}},
}

@inproceedings{choi2018blocking,
    author={Choi, Jee and Liu, Xing and Smith, Shaden and Simon, Tyler},
    booktitle={2018 IEEE International Parallel and Distributed Processing Symposium (IPDPS)}, 
    title={{Blocking Optimization Techniques for Sparse Tensor Computation}}, 
    year={2018},
    publisher = {IEEE},
    address = {USA},
    volume={},
    number={},
    pages={568--577},
    note={{DOI}: \href{https://doi.org/10.1109/IPDPS.2018.00066}{10.1109/IPDPS.2018.00066}},
}

@inproceedings{smith2017accelerating,
    author="Smith, Shaden and Karypis, George",
    editor="Rivera, Francisco F. and Pena, Tom{\'a}s F. and Cabaleiro, Jos{\'e} C.",
    title={{Accelerating the Tucker Decomposition with Compressed Sparse Tensors}},
    booktitle="Euro-Par 2017: Parallel Processing",
    year="2017",
    publisher="Springer International Publishing",
    address="Cham",
    pages="653--668",
    isbn="978-3-319-64203-1",
    note={{DOI}: \href{https://doi.org/10.1007/978-3-319-64203-1_47}{10.1007/978-3-319-64203-1\_47}}
}

@inproceedings{zhao2018bridging,
    author = {Zhao, Yue and Li, Jiajia and Liao, Chunhua and Shen, Xipeng},
    title = {{Bridging the Gap between Deep Learning and Sparse Matrix Format Selection}},
    year = {2018},
    isbn = {9781450349826},
    publisher = {Association for Computing Machinery},
    address = {New York, NY, USA},
    note = {{DOI}: \href{https://doi.org/10.1145/3178487.3178495}{10.1145/3178487.3178495}},
    booktitle = {Proceedings of the 23rd ACM SIGPLAN Symposium on Principles and Practice of Parallel Programming},
    pages = {94--108},
    numpages = {15},
    location = {Vienna, Austria},
    series = {PPoPP '18}
}

@inproceedings{xie2019ia,
    author = {Xie, Zhen and Tan, Guangming and Liu, Weifeng and Sun, Ninghui},
    title = {{IA-SpGEMM: an Input-Aware Auto-Tuning Framework for Parallel Sparse Matrix-Matrix Multiplication}},
    year = {2019},
    isbn = {9781450360791},
    publisher = {Association for Computing Machinery},
    address = {New York, NY, USA},
    note = {{DOI}: \href{https://doi.org/10.1145/3330345.3330354}{10.1145/3330345.3330354}},
    booktitle = {Proceedings of the ACM International Conference on Supercomputing},
    pages = {94--105},
    numpages = {12},
    location = {Phoenix, Arizona},
    series = {ICS '19}
}

@article{KoBa09,
    author = {Kolda, Tamara G. and Bader, Brett W.},
    title = {{Tensor Decompositions and Applications}},
    journal = {SIAM Review},
    volume = {51},
    number = {3},
    pages = {455--500},
    year = {2009},
    note = {{DOI}: \href{https://doi.org/10.1137/07070111X}{10.1137/07070111X}},
}

@inproceedings{Smith2015a,
    author = {Smith, Shaden and Karypis, George},
    title = {{Tensor-Matrix Products with a Compressed Sparse Tensor}},
    year = {2015},
    isbn = {9781450340014},
    publisher = {Association for Computing Machinery},
    address = {New York, NY, USA},
    note = {{DOI}: \href{https://doi.org/10.1145/2833179.2833183}{10.1145/2833179.2833183}},
    booktitle = {Proceedings of the 5th Workshop on Irregular Applications: Architectures and Algorithms},
    articleno = {5},
    pages={1--7},
    numpages = {7},
    location = {Austin, Texas},
    series = {$\textrm{IA}^3$ '15}
}

@INPROCEEDINGS{Smith2015,
    author={Smith, Shaden and Ravindran, Niranjay and Sidiropoulos, Nicholas D. and Karypis, George},
    booktitle={2015 IEEE International Parallel and Distributed Processing Symposium}, 
    title={{SPLATT: Efficient and Parallel Sparse Tensor-Matrix Multiplication}}, 
    year={2015},
    volume={},
    number={},
    pages={61--70},
    address = {USA},
	publisher = {IEEE},
    note={{DOI}: \href{https://doi.org/10.1109/IPDPS.2015.27}{10.1109/IPDPS.2015.27}}
}

@article{KR_product,
    author = {Shangzhi Liu AND G\"otz Trenkler},
    title = {{Hadamard, Khatri-Rao, Kronecker, and Other Matrix Products}},
    journal = {International Journal of Information and Systems Sciences},
    year = {2008},
    pages = {160--177},
    volume = {4},
    number = {1}
}

@INPROCEEDINGS{alto_stream_2023,
    author={Soh, Yongseok and Helal, Ahmed E. and Checconi, Fabio and Laukemann, Jan and Tithi, Jesmin Jahan and Ranadive, Teresa and Petrini, Fabrizio and Choi, Jee W.},
    booktitle={2023 IEEE International Parallel and Distributed Processing Symposium (IPDPS)}, 
    title={{Dynamic Tensor Linearization and Time Slicing for Efficient Factorization of Infinite Data Streams}}, 
    year={2023},
    volume={},
    number={},
    pages={402--412},
    address = {USA},
	publisher = {IEEE},
    note={{DOI}: \href{https://doi.org/10.1109/IPDPS54959.2023.00048}{10.1109/IPDPS54959.2023.00048}}
}

@inproceedings{kurt2022sparsity,
    author={Kurt, Süreyya Emre and Raje, Saurabh and Sukumaran-Rajam, Aravind and Sadayappan, P.},
    booktitle={2022 IEEE International Parallel and Distributed Processing Symposium (IPDPS)}, 
    title={{Sparsity-Aware Tensor Decomposition}}, 
    year={2022},
    volume={},
    number={},
    pages={952--962},
    address = {USA},
	publisher = {IEEE},
    note={{DOI}: \href{https://doi.org/10.1109/IPDPS53621.2022.00097}{10.1109/IPDPS53621.2022.00097}}
}

@inproceedings{li2017model,
    author={Li, Jiajia and Choi, Jee and Perros, Ioakeim and Sun, Jimeng and Vuduc, Richard},
    booktitle={2017 IEEE International Parallel and Distributed Processing Symposium (IPDPS)}, 
    title={{Model-Driven Sparse CP Decomposition for Higher-Order Tensors}}, 
    year={2017},
    volume={},
    number={},
    pages={1048--1057},
    address = {USA},
	publisher = {IEEE},
    note={{DOI}: \href{https://doi.org/10.1109/IPDPS.2017.80}{10.1109/IPDPS.2017.80}}
}

@book{stanley2011enumerative,
  title={{Enumerative Combinatorics, Second Edition}},
  author={Stanley, Richard P},
  volume={1},
  series={Cambridge Studies in Advanced Mathematics},
  year={2011},
  isbn={9781107602625},
  publisher = {Cambridge University Press},
  address = {UK},
  note={{DOI}: \href{https://doi.org/10.1017/CBO9781139058520}{10.1017/CBO9781139058520}},
}

@article{mankowitz2023faster,
  title={{Faster Sorting Algorithms Discovered Using Deep Reinforcement Learning}},
  author={Mankowitz, Daniel J and Michi, Andrea and Zhernov, Anton and Gelmi, Marco and Selvi, Marco and Paduraru, Cosmin and Leurent, Edouard and Iqbal, Shariq and Lespiau, Jean-Baptiste and Ahern, Alex and others},
  journal={Nature},
  volume={618},
  number={7964},
  pages={257--263},
  year={2023},
  publisher={Nature Publishing Group UK London},
  note={{DOI}: \href{https://doi.org/10.1038/s41586-023-06004-9}{10.1038/s41586-023-06004-9}}
}

@article{fawzi2022discovering,
  title={{Discovering Faster Matrix Multiplication Algorithms with Reinforcement Learning}},
  author={Fawzi, Alhussein and Balog, Matej and Huang, Aja and Hubert, Thomas and Romera-Paredes, Bernardino and Barekatain, Mohammadamin and Novikov, Alexander and R. Ruiz, Francisco J and Schrittwieser, Julian and Swirszcz, Grzegorz and others},
  journal={Nature},
  volume={610},
  number={7930},
  pages={47--53},
  year={2022},
  publisher={Nature Publishing Group UK London},
  note={{DOI}: \href{https://doi.org/10.1038/s41586-022-05172-4}{10.1038/s41586-022-05172-4}}
}

@misc{ahn2019reinforcement,
      title={{Reinforcement Learning and Adaptive Sampling for Optimized DNN Compilation}}, 
      author={Byung Hoon Ahn and Prannoy Pilligundla and Hadi Esmaeilzadeh},
      year={2019},
      eprint={1905.12799},
      archivePrefix={arXiv},
      primaryClass={cs.LG},
      note={{DOI}: \href{https://doi.org/10.48550/arXiv.1905.12799}{10.48550/arXiv.1905.12799}},
}

@inproceedings{mammadli2020static,
  title={{Static Neural Compiler Optimization via Deep Reinforcement Learning}},
  author={Mammadli, Rahim and Jannesari, Ali and Wolf, Felix},
  booktitle={2020 IEEE/ACM 6th Workshop on the LLVM Compiler Infrastructure in HPC (LLVM-HPC) and Workshop on Hierarchical Parallelism for Exascale Computing (HiPar)},
  pages={1--11},
  year={2020},
  address = {USA},
  publisher = {IEEE},
  note={{DOI}: \href{https://doi.org/10.1109/LLVMHPCHiPar51896.2020.00006}{10.1109/LLVMHPCHiPar51896.2020.00006}},
}

@article{mangalampalli2024efficient,
  title={{Efficient Deep Reinforcement Learning Based Task Scheduler in Multi Cloud Environment}},
  author={Mangalampalli, Sudheer and Karri, Ganesh Reddy and Ratnamani, MV and Mohanty, Sachi Nandan and Jabr, Bander A and Ali, Yasser A and Ali, Shahid and Abdullaeva, Barno Sayfutdinovna},
  journal={Scientific Reports},
  volume={14},
  number={1},
  pages={21850},
  year={2024},
  publisher={Nature Publishing Group UK London},
  note={{DOI}: \href{https://doi.org/10.1038/s41598-024-72774-5}{10.1038/s41598-024-72774-5}}
}

@misc{gu2025deep,
      title={{Deep Reinforcement Learning for Job Scheduling and Resource Management in Cloud Computing: An Algorithm-Level Review}}, 
      author={Yan Gu and Zhaoze Liu and Shuhong Dai and Cong Liu and Ying Wang and Shen Wang and Georgios Theodoropoulos and Long Cheng},
      year={2025},
      eprint={2501.01007},
      archivePrefix={arXiv},
      primaryClass={cs.DC},
      note={{DOI}: \href{https://doi.org/10.48550/arXiv.2501.01007}{10.48550/arXiv.2501.01007}},
}

@inproceedings{armstrong2008reinforcement,
  title={{Reinforcement Learning for Automated Performance Tuning: Initial Evaluation for Sparse Matrix Format Selection}},
  author={Armstrong, Warren and Rendell, Alistair P},
  booktitle={2008 IEEE International Conference on Cluster Computing},
  pages={411--420},
  year={2008},
  publisher={IEEE},
  address = {USA},
  note={{DOI}: \href{https://doi.org/10.1109/CLUSTR.2008.4663802}{10.1109/CLUSTR.2008.4663802}}
}

@article{chen2023accelerating,
  title={{Accelerating Matrix-Centric Graph Processing on GPUs through Bit-Level Optimizations}},
  author={Chen, Jou-An and Sung, Hsin-Hsuan and Shen, Xipeng and Tallent, Nathan and Barker, Kevin and Li, Ang},
  journal={Journal of Parallel and Distributed Computing},
  volume={177},
  pages={53--67},
  year={2023},
  publisher={Elsevier},
  issn = {0743-7315},
  note = {{DOI}: \href{https://doi.org/10.1016/j.jpdc.2023.02.013}{10.1016/j.jpdc.2023.02.013}},
}

@inproceedings{stylianou2023optimizing,
  title={{Optimizing Sparse Linear Algebra through Automatic Format Selection and Machine Learning}},
  author={Stylianou, Christodoulos and Weiland, Michele},
  booktitle={2023 IEEE International Parallel and Distributed Processing Symposium Workshops (IPDPSW)},
  pages={734--743},
  year={2023},
  address = {USA},
  publisher = {IEEE},
  note={{DOI}: \href{https://doi.org/10.1109/IPDPSW59300.2023.00125}{10.1109/IPDPSW59300.2023.00125}},
}

@article{xiao2024machine,
  title={{Machine Learning-Based Kernel Selector for SpMV Optimization in Graph Analysis}},
  author={Xiao, Guoqing and Zhou, Tao and Chen, Yuedan and Hu, Yikun and Li, Kenli},
  journal={ACM Transactions on Parallel Computing},
  volume={11},
  number={2},
  pages={1--25},
  year={2024},
  publisher={ACM New York, NY},
  note={{DOI}: \href{https://doi.org/10.1145/3652579}{10.1145/3652579}},
}

@inproceedings{yesil2022dense,
  title={{Dense Dynamic Blocks: Optimizing SpMM for Processors with Vector and Matrix Units using Machine Learning Techniques}},
  author={Yesil, Serif and Moreira, Jos{\'e} E and Torrellas, Josep},
  booktitle={Proceedings of the 36th ACM International Conference on Supercomputing},
  pages={1--14},
  address = {USA},
  publisher = {ACM},
  year={2022},
  note={{DOI}: \href{https://doi.org/10.1145/3524059.3532369}{10.1145/3524059.3532369}}
}

@article{gao2024revisiting,
  title={{Revisiting Thread Configuration of SpMV Kernels on GPU: A Machine Learning Based Approach}},
  author={Gao, Jianhua and Ji, Weixing and Liu, Jie and Wang, Yizhuo and Shi, Feng},
  journal={Journal of Parallel and Distributed Computing},
  volume={185},
  pages={104799},
  year={2024},
  publisher={Elsevier},
  issn = {0743-7315},
  note = {{DOI}: \href{https://doi.org/10.1016/j.jpdc.2023.104799}{10.1016/j.jpdc.2023.104799}},  
}

@ARTICLE{liu2023tensor,
  author={Liu, Xingyi and Parhi, Keshab K.},
  journal={IEEE Circuits and Systems Magazine}, 
  title={{Tensor Decomposition for Model Reduction in Neural Networks: A Review [Feature]}}, 
  year={2023},
  volume={23},
  number={2},
  pages={8--28},
  note={{DOI}: \href{https://doi.org/10.1109/MCAS.2023.3267921}{10.1109/MCAS.2023.3267921}},
}

@article{de2008tensor,
  title={{Tensor Rank and the Ill-Posedness of the Best Low-Rank Approximation Problem}},
  author={De Silva, Vin and Lim, Lek-Heng},
  journal={SIAM Journal on Matrix Analysis and Applications},
  volume={30},
  number={3},
  pages={1084--1127},
  year={2008},
  note = {{DOI}: \href{https://doi.org/10.1137/06066518X}{10.1137/06066518X}},  
  publisher={SIAM}
}

@article{kaya2018trees,
author = {Kaya, Oguz and U\c{c}ar, Bora},
title = {{Parallel Candecomp/Parafac Decomposition of Sparse Tensors Using Dimension Trees}},
journal = {SIAM Journal on Scientific Computing},
volume = {40},
number = {1},
pages = {C99--C130},
year = {2018},
note = {{DOI}: \href{https://doi.org/10.1137/16M1102744}{10.1137/16M1102744}},
}

@article{10.1145/3519383,
author = {Psarras, Christos and Karlsson, Lars and Bro, Rasmus and Bientinesi, Paolo},
title = {{Algorithm~1026: Concurrent Alternating Least Squares for Multiple Simultaneous Canonical Polyadic Decompositions}},
year = {2022},
issue_date = {September 2022},
publisher = {Association for Computing Machinery},
address = {New York, NY, USA},
volume = {48},
number = {3},
issn = {0098-3500},
note = {{DOI}: \href{https://doi.org/10.1145/3519383}{10.1145/3519383}},
journal = {ACM Trans. Math. Softw.},
month = sep,
articleno = {34},
numpages = {20},
pages={1--20}
}

@article{10.1002/nla.2511,
author = {Minster, Rachel and Viviano, Irina and Liu, Xiaotian and Ballard, Grey},
title = {{CP Decomposition for Tensors via Alternating Least Squares with QR Decomposition}},
journal = {Numerical Linear Algebra with Applications},
volume = {30},
number = {6},
pages = {e2511},
note = {{DOI}: \href{https://doi.org/10.1002/nla.2511}{10.1002/nla.2511}},
year = {2023}
}

@article{kjolstad:2017:taco,
  author = {Kjolstad, Fredrik and Kamil, Shoaib and Chou, Stephen and Lugato, David and Amarasinghe, Saman},
  title = {{The Tensor Algebra Compiler}},
  journal = {Proc. ACM Program. Lang.},
  issue_date = {October 2017},
  volume = {1},
  number = {OOPSLA},
  month = oct,
  year = {2017},
  issn = {2475-1421},
  pages = {77:1--77:29},
  articleno = {77},
  numpages = {29},
  note = {{DOI}: \href{https://doi.org/10.1145/3133901}{10.1145/3133901}},
  acmid = {3133901},
  publisher = {ACM},
  address = {New York, NY, USA}
}

@INPROCEEDINGS{10818187,
  author={Qu, Xiangjun and Gong, Lei and Lou, Wenqi and Cheng, Qianyu and Chen, Xianglan and Wang, Chao and Zhou, Xuehai},
  booktitle={2024 IEEE 42nd International Conference on Computer Design (ICCD)}, 
  title={{AutoSparse: A Source-to-Source Format and Schedule Auto-Tuning Framework for Sparse Tensor Program}}, 
  year={2024},
  volume={},
  number={},
  pages={504--512},
  note={{DOI}: \href{https://doi.org/10.1109/ICCD63220.2024.00083}{10.1109/ICCD63220.2024.00083}},
  }

@inproceedings{10.5555/645528.657613,
author = {Ng, Andrew Y. and Harada, Daishi and Russell, Stuart J.},
title = {{Policy Invariance Under Reward Transformations: Theory and Application to Reward Shaping}},
year = {1999},
isbn = {1558606122},
publisher = {Morgan Kaufmann Publishers Inc.},
address = {San Francisco, CA, USA},
booktitle = {Proceedings of the Sixteenth International Conference on Machine Learning},
pages = {278--287},
numpages = {10},
series = {ICML '99},
note = {{DOI}: \href{https://dl.acm.org/doi/10.5555/645528.657613}{10.5555/645528.657613}},
}

\end{document}